\documentclass[final]{cvpr}

\usepackage{times}
\usepackage{epsfig}
\usepackage{graphicx}
\usepackage{amsmath}
\usepackage{amssymb}

\usepackage{xcolor}

\usepackage{graphicx}
\usepackage{comment}
\usepackage{amsmath,amssymb} % define this before the line numbering.
\usepackage{color, colortbl}

\usepackage{xcolor}
\usepackage{nicefrac}
\usepackage{booktabs}
\usepackage{placeins}
\usepackage{pifont}
\usepackage{subcaption}
\usepackage{xspace}
\usepackage{arydshln}
\usepackage{nicefrac}

\usepackage{algorithm}
\usepackage{algpseudocode}
\algrenewcommand\algorithmicindent{1.0em}%

\usepackage{array}
\usepackage{booktabs}
\usepackage{adjustbox}

\usepackage{listings}

\usepackage[colorlinks,citecolor=red,urlcolor=blue,bookmarks=false,hypertexnames=true,pagebackref=true,breaklinks=true]{hyperref}

\def\cvprPaperID{448} % *** Enter the CVPR Paper ID here
\def\confYear{CVPR 2021}

\definecolor{BrickRed}{HTML}{B6321C}
\definecolor{RoyalBlue}{HTML}{0071BC}
\definecolor{PineGreen}{HTML}{008B72}
\definecolor{bluefig}{HTML}{5B9BD5}
\definecolor{Gray}{gray}{0.9}

\newcommand{\tbd}{\textcolor{BrickRed}{TBD}}

\newcommand{\OK}{\ding{51}}
\let\thetaold\theta
\renewcommand{\theta}{\boldsymbol{\thetaold}}
\newcommand{\mcL}{\mathcal{L}}
\newcommand{\vx}{\mathbf{x}}
\newcommand{\vh}{\mathbf{h}}

\newcommand{\vp}{\mathbf{p}}
\DeclareMathOperator*{\argmax}{argmax} % thin space, limits underneath in displays
\newcommand{\pp}{\,\textit{p.p}}

\newcommand{\tableindent}{\,\,\,}
\newcommand{\ours}{PLOP\,}

\begin{document}

%%%%%%%%% TITLE
\title{PLOP: Learning without Forgetting for Continual Semantic Segmentation}
% PLOP: Pseudo-labeling and LOcal Pooled Output Distillation for Continual Semantic Segmentation

\author{Arthur Douillard\textsuperscript{1,2}, Yifu Chen\textsuperscript{1}, Arnaud Dapogny\textsuperscript{3}, Matthieu Cord\textsuperscript{1,4}\\
\textsuperscript{1}Sorbonne Université, \textsuperscript{2}Heuritech, \textsuperscript{3}Datakalab, \textsuperscript{4}valeo.ai
\\{\tt\small arthur.douillard@heuritech.com, \{yifu.chen, matthieu.cord\}@lip6.fr, ad@datakalab.com}
}

\maketitle

\begin{abstract}
   Deep learning approaches are nowadays ubiquitously used to tackle computer vision tasks such as semantic segmentation, requiring large datasets and substantial computational power. Continual learning for semantic segmentation (CSS) is an emerging trend that consists in updating an old model by sequentially adding new classes. However, continual learning methods are usually prone to catastrophic forgetting.
   This issue is further aggravated in CSS where, at each step, old classes from previous iterations are collapsed into the background. In this paper, we propose Local POD, a multi-scale pooling distillation scheme that preserves long- and short-range spatial relationships at feature level. Furthermore, we design an entropy-based pseudo-labelling of the background w.r.t. classes predicted by the old model to deal with background shift and avoid catastrophic forgetting of the old classes. Our approach, called PLOP, significantly outperforms state-of-the-art methods in existing CSS scenarios, as well as in newly proposed challenging benchmarks\footnote{Code is available at \\\url{https://github.com/arthurdouillard/CVPR2021_PLOP}}.
\end{abstract}
\section{Introduction}

\begin{figure}
\centering
  \includegraphics[width=1.0\linewidth]{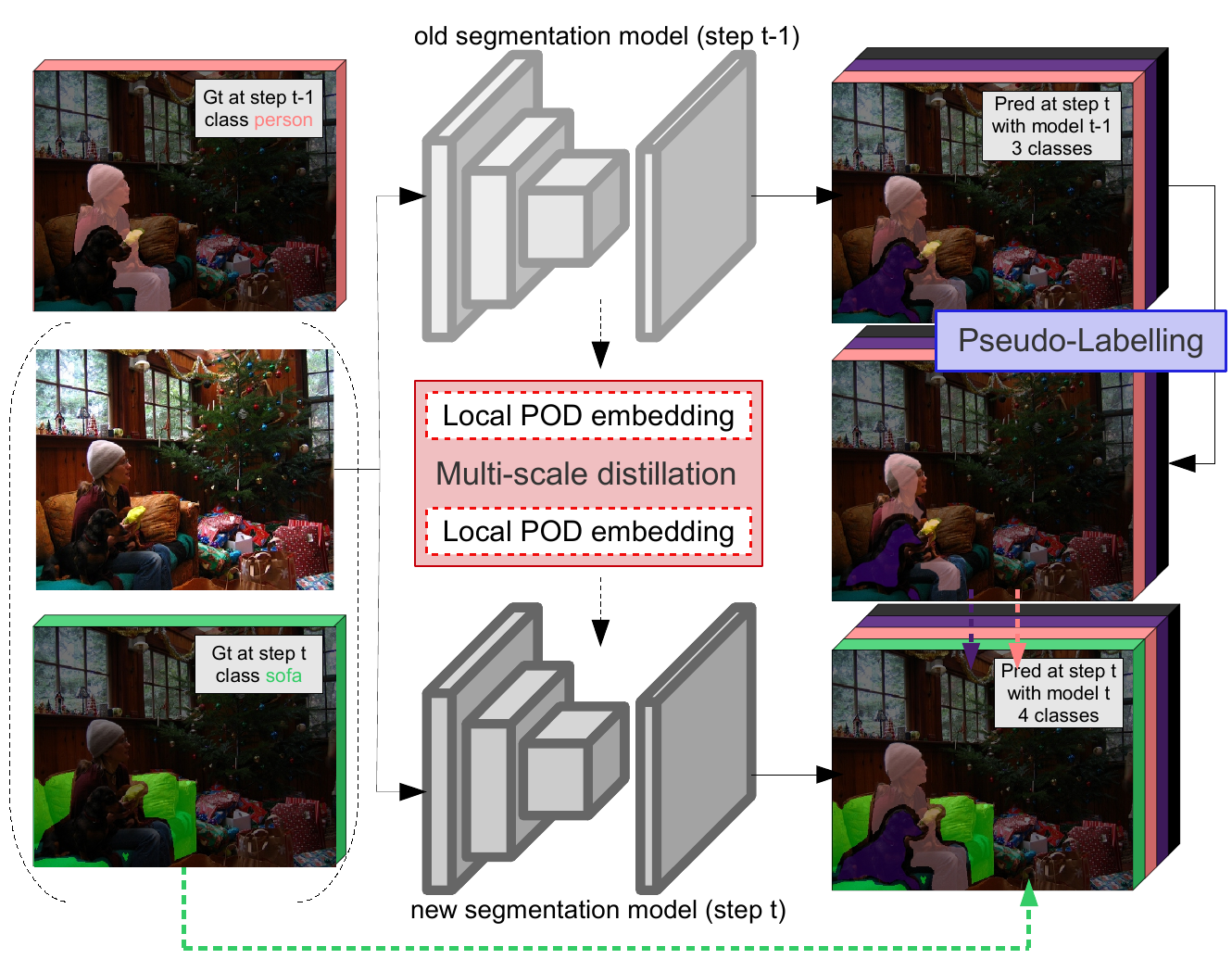}
  \caption{Our two-part strategy aims at learning a segmentation network in a continual learning framework, where old class pixels are collapsed into the background at current stage. We generate pseudo labels from old predictions ({\color{blue}blue}) to deal with the background shift, and retain short- and long-range spatial dependencies by Local POD distillation ({\color{red}red}) to prevent catastrophic forgetting.}
\label{fig:bigpicture}
\end{figure}

Semantic segmentation is a fundamental problem of computer vision, that aims at assigning a label to each pixel of an image. In recent years, the introduction of Convolutional Neural Networks (CNNs) has addressed semantic segmentation in a traditional framework, where all classes are known beforehand and learned at once ~\cite{tao2020HRNet,zhang2020resnest,chen2018ZPSA}. This setup, however, is quite limited for practical applications. In a more realistic scenario, the model should be able to continuously learn new classes without retraining from scratch. This setup, referred here as Continual Semantic Segmentation (CSS), has emerged very recently for medical applications \cite{ozdemir2018learnthenewkeeptheold,ozdemir2019segmentationanotomical} before being proposed for general segmentation datasets \cite{michieli2019ilt,cermelli2020modelingthebackground}. 

Deep learning approaches that deal with CSS face two main challenges. The first one, inherited from continual learning, is called \textit{catastrophic forgetting}~\cite{robins1995catastrophicforgetting,french1999catastrophicforgetting,thrun1998lifelonglearning}, and points to the fact that neural networks tend to completely and abruptly forget previously learned knowledge when learning new information ~\cite{kemker2018measuringforgetting}. Catastrophic forgetting presents a real challenge for continual learning applications based on deep learning methods, especially when storing previously seen data is not allowed for privacy reasons.

The second issue, CSS specific, is the semantic shift of the background class. In a traditional semantic segmentation setup, the background contains pixels that don't belong to any other class. However, in CSS, the background contains pixels that don't belong to any of the \textit{current} classes. Thus, for a specific learning step, the background can contain both future classes, not yet seen by the model, as well as old classes. Thus, if nothing is done to distinguish pixels belonging to the real background class from old class pixels, this background shift phenomenon risks exacerbating the catastrophic forgetting even further \cite{cermelli2020modelingthebackground}.

In this paper, we propose a deep learning strategy to address these two challenges in CSS. Instead of reusing old images, our approach, called \ours, standing for Pseudo-label and LOcal POD leverages the old model in two manners, as illustrated on \autoref{fig:bigpicture}. First, we propose a feature-based multi-scale distillation scheme to alleviate catastrophic forgetting. Second, we employ a confidence-based pseudo-labeling strategy to retrieve old class pixels within the background. For instance, if a current ground truth mask only distinguish pixels from class \texttt{sofa} and background, our approach allows to assign old classes to background pixels, e.g. classes \texttt{person}, \texttt{dog} or \texttt{background} (the semantic class).

We thoroughly validate \ours on several datasets, showcasing significant performance improvements compared to the state-of-the-art methods in existing CSS scenarios. Furthermore, we propose several novel scenarios to further quantify the performances of CSS methods when it comes to long term learning, class presentation order and domain shift. Last but not least, we show that \ours largely outperforms every CSS approach in these scenarios. To sum it up, our contributions are three-folds:

\begin{itemize}
    \itemsep-0.4em 
    \item We propose a multi-scale spatial distillation loss to better retain knowledge through the continual learning steps, by preserving long- and short-range spatial relationships, avoiding catastrophic forgetting.
    \item We introduce a confidence-based pseudo-labeling strategy to identify old classes for the current background pixels and deal with background shift.
    \item We show that \ours significantly outperforms state-of-the-art approaches in existing scenarios and datasets for CSS, as well as in several newly proposed challenging benchmarks.
\end{itemize}

\section{Related Work}

CSS is a relatively new field where only a few recent papers addressed this specific problem. We thus start this section with a brief overview of the recent advances in semantic segmentation as well as continual learning and follow with a more in-depth discussion of existing approaches to CSS.

\noindent\textbf{Semantic Segmentation} methods based on Fully Convolutional Networks (FCN)  \cite{long2015fcn,sermanet2014overfeat} have achieved impressive results on several segmentation benchmarks ~\cite{everingham2015pascalvoc, cordts2016cityscapes,zhou2017adedataset,caesar2018cocoostuff}. These methods improve the segmentation accuracy by incorporating more spatial information or exploiting contextual information specifically. Atrous convolution~\cite{chen2018deeplab,mehta2018espnet} and encoder-decoder architecture~\cite{ronneberger2015UNet,noh2015deconvolution,badrinarayanan2017segnet} are the most common methods for retaining spatial information. Examples of recent works exploiting contextual information include attention mechanisms~\cite{yuan2018ocnet,zhao2018psanet,fu2019DANet,huang2019CCNet,yuan2020ocr,tao2020HRNet,zhang2020resnest}, and fixed-scale aggregation ~\cite{zhao2017PSPNet,chen2018deeplab,chen2018ZPSA,zhang2018ContextEncoding}. More recently, Strip Pooling~\cite{hou2020strippooling} consists in pooling along the width or height dimensions similarly to POD \cite{douillard2020podnet} as a complement to a spatial pyramid pooling \cite{he2014spatialpyramidpooling} to capture both global and local statistics.

\noindent\textbf{Continual Learning} models generally face the challenge of catastrophic forgetting of the old classes \cite{robins1995catastrophicforgetting,thrun1998lifelonglearning,french1999catastrophicforgetting}. Several solutions exist to address this problem: for instance, rehearsal learning consists in keeping a limited amount of training data from old classes either as raw images~\cite{robins1995catastrophicforgetting,rebuffi2017icarl,castro2018end_to_end_inc_learn,chaudhry2019tinyepisodicmemories}, compressed features~\cite{hayes2020remind,iscen2020incrementalfeatureadaptation}, or generated training data~\cite{kemker2018fearnet,shin2017deep_generative_replay,liu2020mnemonics}.
Other works focus on adaptive architectures that can extend themselves to integrate new classes~\cite{yoon2018dynamically_expandable_networks,li2019learning_to_grow} or dynamically re-arrange co-existing sub-networks \cite{frankle2019lottery_ticket} each specialized in one specific task~\cite{fernando2017path_net,golkar2019neural_pruning, hung2019cpg}, or to explicitly correct the classifier drift ~\cite{wu2019bias_correction,zhao2020weightalignement,belouadah2019il2m,belouadah2020scail} that happens with continually changing class distributions. Last but not least, distillation-based methods aim at constraining the model as it changes, either directly on the weights~\cite{kirkpatrick2017ewc,aljundi2018MemoryAwareSynapses,chaudhry2018riemannien_walk,zenke2017synaptic_intelligence}, the gradients~\cite{lopezpaz2017gem,chaudhry2019AGEM}, the output probabilities~\cite{li2018lwf,rebuffi2017icarl,castro2018end_to_end_inc_learn,cermelli2020modelingthebackground}, intermediary features~\cite{hou2019ucir,dhar2019learning_without_memorizing_gradcam,peng2019m2kd,douillard2020podnet}, or combinations thereof.

\noindent\textbf{Continual Semantic segmentation}: Despite enormous progress in the two aforementioned areas respectively, segmentation algorithms are mostly used in an offline setting, while continual learning methods generally focus on image classification. Recent works extend existing continual learning methods \cite{li2018lwf,hou2019ucir} for medical applications \cite{ozdemir2018learnthenewkeeptheold,ozdemir2019segmentationanotomical} and general 
semantic segmentation \cite{michieli2019ilt}.
The latter considers that the previously learned categories are properly annotated in the images of the new dataset. This is an unrealistic assumption that fails to consider the background shift: pixels labeled as background at the current step are semantically ambiguous, in that they can contain pixels from old classes (including the real semantic background class, which is generally deciphered first) as well as pixels from future classes. To the best of our knowledge, Cermelli et al.~\cite{cermelli2020modelingthebackground} are the first to address this background shift problem along with catastrophic forgetting. 
To do so, they apply two loss terms at the output level. First, they use a knowledge distillation loss to reduce forgetting. However, only constraining the output of the network with a distillation term is not enough to preserve the knowledge of the old classes, leading to too much plasticity and, ultimately, catastrophic forgetting. Second, they propose to modify the traditional cross-entropy loss for background pixels to propagate only the sum probability of old classes throughout the continual learning steps. We argue that this constraint is not strong enough to preserve a high discriminative power w.r.t. the old classes when learning new classes under background shift. On the contrary, in what follows, we introduce our \ours framework and show how it enables learning without forgetting for CSS.

\section{PLOP Segmentation Learning Framework}

\subsection{Continual semantic segmentation framework}\label{overview}

CSS aims at learning a model in $t=1 \dots T$ steps. For each step, we present a dataset $\mathcal{D}_t$ that consists in a set of pairs $(I^t, S^t)$, where $I^t$ denotes an input image of size $W \times H$ and $S^t$ the corresponding ground truth segmentation mask. The latter only contains the labels of current classes $\mathcal{C}^t$, and all other labels (e.g. old classes $\mathcal{C}^{1:t-1}$ or future classes $\mathcal{C}^{t+1:T}$) are collapsed into the background class $c_\text{bg}$. However, the model at step $t$ shall be able to predict all the classes seen over time $\mathcal{C}^{1:t}$. Consequently, we identify two major pitfalls in CSS: the first one, catastrophic forgetting \cite{robins1995catastrophicforgetting,french1999catastrophicforgetting}, suggests that the network will completely forget the old classes $\mathcal{C}^{1:t-1}$ when learning $\mathcal{C}^t$. Furthermore, catastrophic forgetting is aggravated by the second pitfall, the background shift: at step $t$, the pixels labeled as background are indeed ambiguous, as they may contain either old (including the real background class, predicted in $\mathcal{C}^{1}$) or future classes. \autoref{fig:examplescss} (top row) illustrates background shift.

Classically, a deep model at step $t$ can be written as the composition of a feature extractor $f^t(\cdot)$ and a classifier $g^t(\cdot)$. Features can be extracted at any layer $l$ of the former $f_l^t(\cdot)\,, l \in \{1, ... L\}$. We denote $\hat S^t=g^t \circ f^t(I)$ the output predicted segmentation mask and $\Theta^t$ the set of learnable parameters for the current network at step $t$.

\subsection{Multi-scale local distillation with Local POD}\label{distillation}

A common solution to alleviate catastrophic forgetting in continual learning consists of using a distillation loss between the predictions of the old and current models \cite{li2018lwf}. This distillation loss should constitute a suitable trade-off between too much rigidity (\textit{i.e.} enforcing too strong constraints, resulting in not being able to learn new classes) and too much plasticity (\textit{i.e.} enforcing loose constraints, which leads to catastrophic forgetting of the old classes).

Among existing distillation schemes based on intermediate features \cite{douillard2020podnet,zagoruyko2016distillation_attention,romero2014fitnet_hints,dhar2019learning_without_memorizing_gradcam,peng2019m2kd,hou2019ucir}, POD \cite{douillard2020podnet} consists in matching global statistics at different feature levels between the old and current models. Let $\vx$ denote an embedding tensor of size $H \times W \times C$. Extracting a POD embedding $\Phi$ consists in concatenating the $H \times C$ width-pooled slices and the $W \times C$ height-pooled slices of $\vx$:

\begin{equation}
    \Phi(\vx) = \left[\frac{1}{W} \sum_{w=1}^W \vx[:,w,:] \bigg\Vert \frac{1}{H} \sum_{h=1}^H \vx[h,:,:]\right] \in \mathcal{R}^{(H + W) \times C}\,,
\label{eq:pod_embedding}
\end{equation}

where $[\cdot\,\|\,\cdot]$ denotes concatenation over the channel axis. In our case, this embedding is computed at several layers, for both the old and current model. Then the POD loss consists in minimizing the L2 distance between the two sets of embeddings over the current network parameters $\Theta^t$:

\begin{figure}[ht!]
\centering
  \includegraphics[width=1\linewidth]{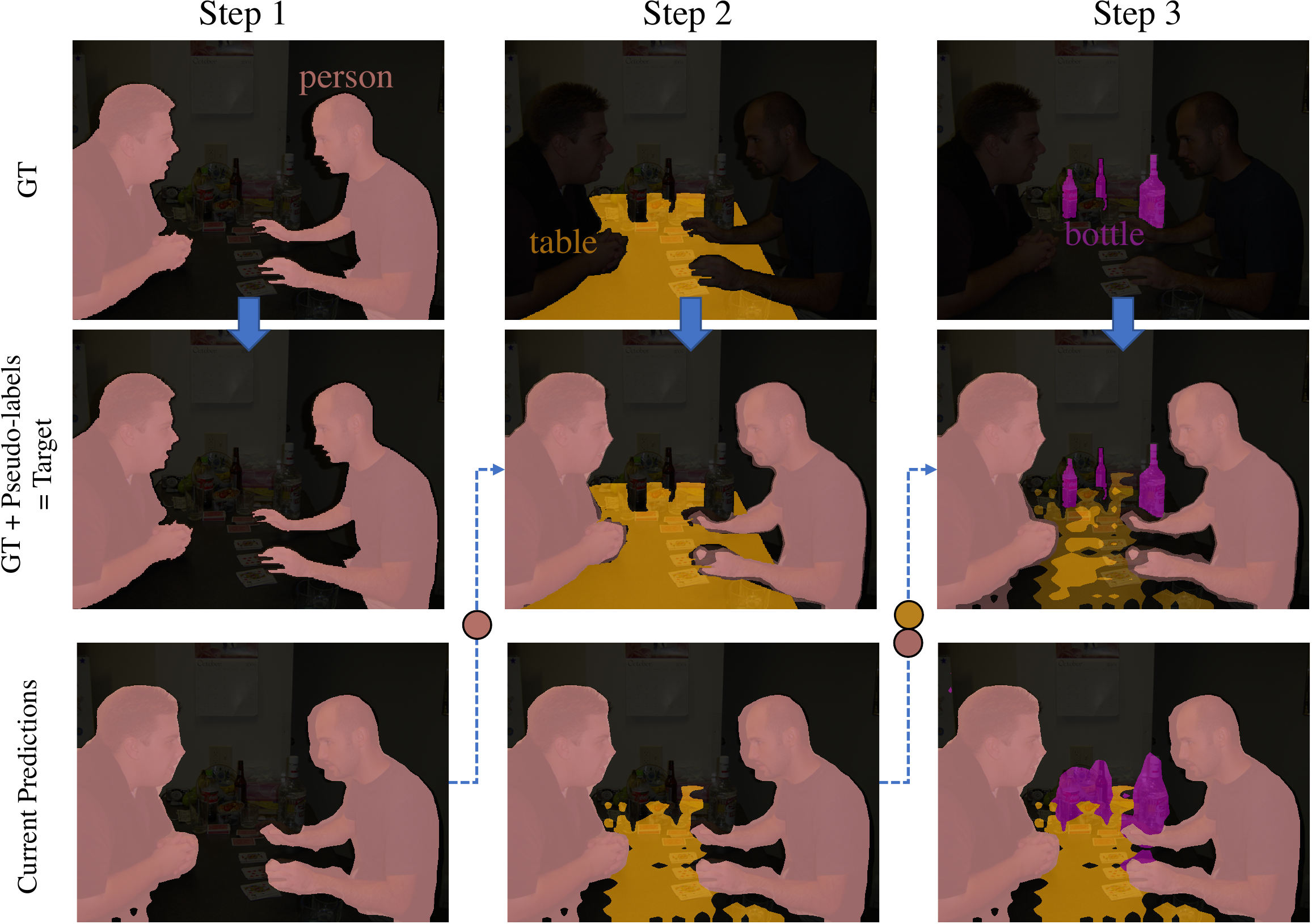}
    \caption{Background shift example in ground truth masks (top row). At step $2$ background pixels contain old ({\color{red}\texttt{person}}) and future classes ({\color{purple}\texttt{bottle}}). The model's target (middle row) is the union of the ground-truth and the pseudo-labels (with transparent filtered uncertain pixels) generated by the previous model. The latter helps the current model predictions (bottom row) to retain information  of the old classes ({\color{orange}\texttt{table}}).}
\label{fig:examplescss}
\end{figure}

\vspace*{-1.35cm}
\begin{equation}
    \mcL_\text{pod}(\Theta^t) = \frac{1}{L} \sum_{l = 1}^L \left\Vert  \Phi(f^t_l(I)) -  \Phi(f^{t-1}_l(I)) \right\Vert^2\,.
\label{eq:pod_loss}
\end{equation}

Due to its ability to constraint spatial statistics instead of raw pixel values, this approach yields state-of-the-art results in the context of continual learning for classification. In the frame of CSS, another interest arises: its ability to model long-range dependencies across a whole axis (horizontal or vertical). However, while spatial information is discarded by global pooling in classification, semantic segmentation requires a higher degree of spatial precision. Therefore, modeling statistics across the whole width or height leads to blurring local statistics important for smaller objects.

Hence, a suitable distillation scheme for CSS shall retain both long-range and short-range spatial relationships. Thus, inspired from the multi-scale literature \cite{lazbnik2006spatial_pyramid_matching,he2014spatialpyramidpooling}, we propose a novel Local POD feature distillation scheme, that consists in computing width and height-pooled slices on multiple regions extracted at different scales $\{1/2^s\}_{s=0 \dots S}$, as shown on \autoref{fig:local_pod}. For an embedding tensor $\vx$ of size $H \times W \times C$, and at scale $1/2^s$, the Local POD embedding $\Psi^s(\vx)$ at scale $s$ is computed as the concatenation of $s^2$ POD embeddings:

\vspace*{-0.55cm}
\begin{equation}
    \Psi^s(\vx) = \left[ \Phi(\vx^s_{0,0}) \| \dots \| \Phi(\vx^s_{s-1,s-1}) \right] \in \mathcal{R}^{(H + W) \times C}\,,
\label{eq:localpod_embedding1}
\end{equation}

where $\forall i = 0 \dots s-1$, $\forall j = 0 \dots s-1$, $\vx^s_{i,j} = \vx[i H/s:(i+1) H/s, j W/s:(j+1) W/s,:]$ is a sub-region of the embedding tensor $\vx$ of size $W/s \times H/s$. We then concatenate (along channel axis) the Local POD embeddings $\Psi^s(\vx)$ of each scale $s$ to form the final embedding:

\begin{figure}[t!]
\centering
  \includegraphics[width=1\linewidth]{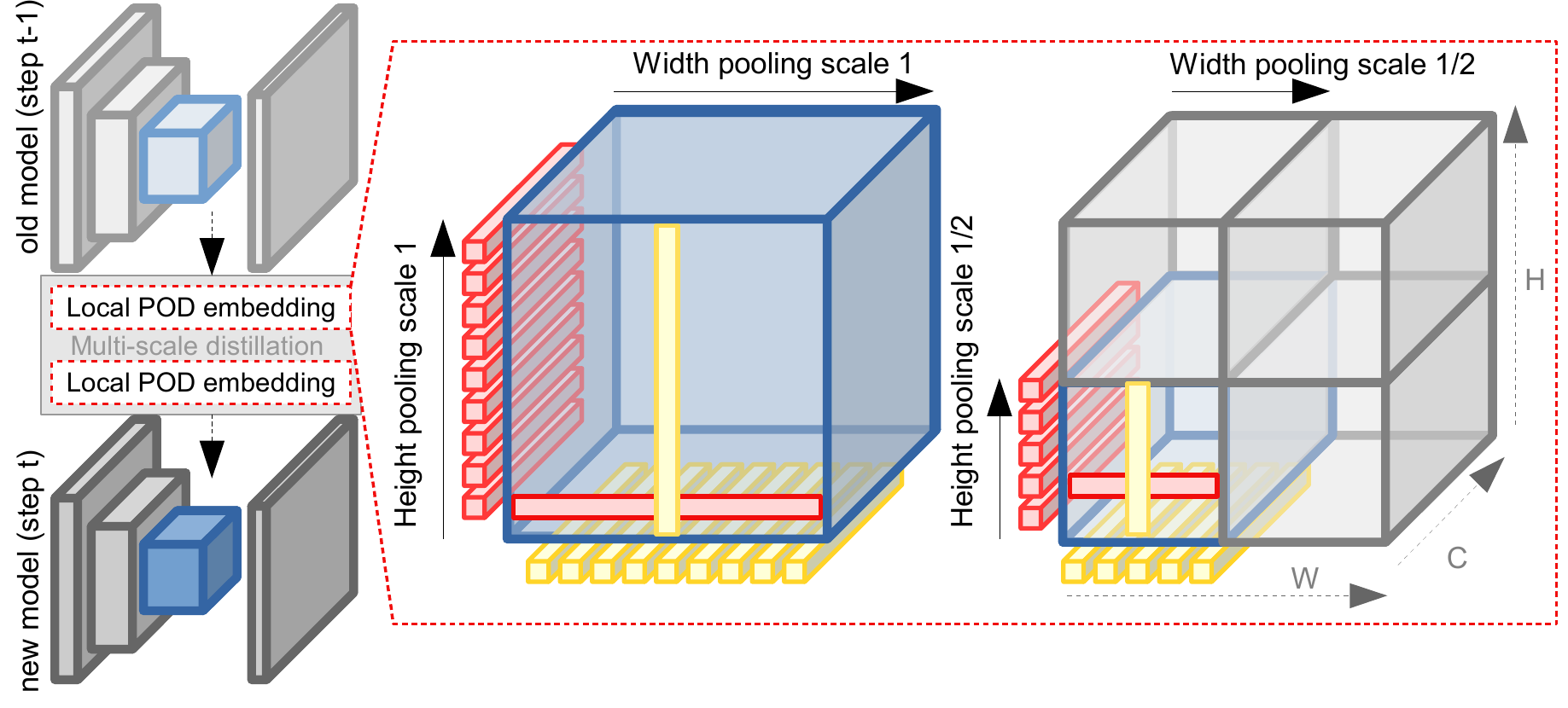} 
    \caption{Illustration of local POD. An embedding of size $W \times H \times C$ is pooled at $S$ scales with POD with a spatial-pyramid scheme. Here applying local POD with $S=2$ and scales 1 and 1/2 respectively produces 1, and 4 POD embeddings making $S \times C \times (H+W)$ dimensions total.}
\label{fig:local_pod}
\end{figure}

\vspace*{-0.5cm}
\begin{equation}
    \Psi(\vx) = \left[ \Psi^1(\vx) \| \dots \| \Psi^S(\vx) \right] \in \mathcal{R}^{(H + W) \times C \times S}\,.
\label{eq:local_pod}
\end{equation}

We provide in the supplementary materials the complete algorithm of Local POD embedding extraction. We compute Local POD embeddings for several layers of both old and current models. The final Local POD loss is:

\vspace*{-0.5cm}
\begin{equation}
    \mcL_{\scriptstyle\text{LocalPod}}(\Theta^t) = \frac{1}{L} \sum_{l = 1}^L \left\Vert  {\Psi}(f^t_l(I)) -  {\Psi}(f^{t-1}_l(I)) \right\Vert^2\,.
\label{eq:local_pod_loss}
\end{equation}

Note that while the first scale of Local POD ($1/2^0$) is equivalent to POD and models long-range dependencies, which are important for segmentation \cite{wang2020axialdeeplab,huang2020ccnet,park2020csc,hou2020strippooling}, the subsequent scales ($s=1/2^1, 1/2^2 \dots$) enforce short-range dependencies. This constrains the old and current models to have similar statistics over more local regions. Thus, Local POD allows retaining both long-range and short-range spatial relationships, thus alleviating catastrophic forgetting.

\subsection{Solving background shift with pseudo-labeling}\label{hardpl}

As described above, the pixels labelled as background at step $t$ can belong to either old (including the semantic background class) or future classes. Thus, treating them as background would result in aggravating catastrophic forgetting. Rather, we address background shift with a pseudo-labeling strategy for background pixels. Pseudo-labeling \cite{lee2013pseudolabel} is commonly used in domain adaptation for semantic segmentation \cite{vu2019advent,li2019bidirectionallearning,zou2018classbalancedselftraining,saporta2020esl}, where a model is trained on the union of real labels of a source dataset and pseudo labels assigned to an unlabeled target dataset. In our case, we use predictions of the old model for background pixels as clues regarding their real class, most notably if they belong to any of the old classes, as illustrated on \autoref{fig:examplescss} (middle row). Formally, let $C^t=card(\mathcal{C}^{t})-1$ the cardinality of the current classes excluding the background class. Let $\hat S^{t} \in \mathcal{R}^{W,H,1+C^1 +\dots + C^t}$ denote the predictions of the current model (which include the real background class, all the old classes as well as the current ones). We define $\tilde S^{t} \in \mathcal{R}^{W,H,1+C^1 +\dots + C^t}$ the target as step $t$, computed using the one-hot ground-truth segmentation map $S^{t} \in \mathcal{R}^{W,H,1+C^t}$ at step $t$ as well as pseudo-labels extracted using the old model predictions $\hat S^{t-1} \in \mathcal{R}^{W,H,1+C^1 +\dots + C^{t-1}}$ as follows:

\vspace*{-0.5cm}
\begin{equation}
\footnotesize
\tilde S^{t}\left(w,h,c\right)= \mkern-5mu \left\{\begin{array}{ll}
\mkern-10mu 1 \mkern-27mu & \text { if } S^{t} (w,h,c_{bg})=0 \text { and } c = \argmax \limits_{c' \in \mathcal{C}^{t}} S^{t}(w,h,c') \\
\mkern-10mu 1 \mkern-27mu & \text { if } S^{t}(w,h,c_{bg})=1 \text { and } c = \mkern-8mu \argmax \limits_{c' \in \mathcal{C}^{1:t-1}} \mkern-6mu \hat S^{t-1}(w,h,c') \\
\mkern-10mu 0 \mkern-27mu & \text { otherwise }\\
\end{array}\right.
\label{eq:pseudo_bis}
\end{equation}

In other words, in the case of non-background pixels we copy the ground truth label. Otherwise, we use the class predicted by the old model $g^{t-1}(f^{t-1}(\cdot))$. This pseudo-label strategy allows to assign each pixel labelled as background his real semantic label if this pixel belongs to any of the old classes. However pseudo-labeling all background pixels can be unproductive, e.g. on uncertain pixels where the old model is likely to fail. Therefore we only retain pseudo-labels where the old model is ``\textit{confident}'' enough. \autoref{eq:pseudo_bis} can be modified to take into account this uncertainty:

\vspace*{-0.5cm}
\begin{equation}
\scriptsize
\tilde S^{t}\left(w,h,c\right)\mkern-5mu = \mkern-5mu \left\{\begin{array}{ll}
\mkern-14mu 1 \mkern-27mu & \mkern-8mu \text { if } S^{t} \mkern-4mu (w,h,c_{bg})\mkern-4mu = \mkern-4mu 0 \text { and } c \mkern-4mu = \mkern-4mu \argmax \limits_{c' \in \mathcal{C}^{t}} S^{t} \mkern-4mu (w,h,c') \\
\mkern-14mu 1 \mkern-27mu & \mkern-8mu \text { if } S^{t} \mkern-4mu (w,h,c_{bg}) \mkern-4mu = \mkern-4mu 1 \text { and } c \mkern-4mu = \mkern-12mu \argmax \limits_{c' \in \mathcal{C}^{1:t-1}} \mkern-6mu \hat S^{t-1} \mkern-4mu (w,h,c') \text{ and } u \mkern-4mu < \mkern-4mu \tau_{c}\\
\mkern-14mu 0 \mkern-27mu & \mkern-4mu \,\text { otherwise\,, }\\
\end{array}\right.
\label{eq:pseudo_bis_uncertain}
\end{equation}

\begin{table*}[t]
\centering
\caption{Continual Semantic Segmentation results on Pascal-VOC 2012 in Mean IoU (\%). $\dagger$: results excerpted from ~\cite{cermelli2020modelingthebackground}. Other results comes from re-implementation.}
\vspace*{-0.3cm}
\label{tab:voc_sota}
\begin{tabular}{@{}l|cccc||cccc||cccc@{}}
\toprule
& \multicolumn{4}{c}{\textbf{19-1} (2 tasks)} & \multicolumn{4}{c}{\textbf{15-5} (2 tasks)} & \multicolumn{4}{c}{\textbf{15-1} (6 tasks)}\\
\textbf{Method} & 0-19 & 20 & \textit{all} & \textit{avg} & 0-15 & 16-20 & \textit{all} & \textit{avg} & 0-15 & 16-20 & \textit{all} & \textit{avg}\\
\midrule
% from paper MiB
%$\text{Fine Tuning}^\dagger$ & \tableindent 6.80 & 12.90 & \tableindent 7.10 &  & \tableindent 2.10 & 33.10 & \tableindent 9.80 &  & \tableindent 0.20 & \tableindent 1.80 & \tableindent 0.60 & \\
%$\text{PI}^\dagger$ \cite{zenke2017synaptic_intelligence} & \tableindent 7.50 & 14.00 & \tableindent 7.80 &  & \tableindent 1.60 & 33.30 & \tableindent 9.50 &  & \tableindent 0.00 & \tableindent 1.80 & \tableindent 0.50 & \\
$\text{EWC}^\dagger$ \cite{kirkpatrick2017ewc} & 26.90 & 14.00 & 26.30 &  & 24.30 & 35.50 & 27.10 &  & \tableindent 0.30 & \tableindent 4.30 & \tableindent 1.30 &  \\
%$\text{RW}^\dagger$ \cite{chaudhry2018riemannien_walk} & 23.30 & 14.20 & 22.90 &  & 16.60 & 34.90 & 21.20 &  & \tableindent 0.00 & \tableindent 5.20 & \tableindent 1.30 & \\
%$\text{LwF}^\dagger$ \cite{li2018lwf} & 51.20 & \tableindent 8.50 & 49.10 &  & 58.90 & 36.60 & 53.30 &  & \tableindent 1.00 & \tableindent 3.90 & \tableindent 1.80 & \\
$\text{LwF-MC}^\dagger$ \cite{rebuffi2017icarl} & 64.40 & 13.30 & 61.90 &  & 58.10 & 35.00 & 52.30 &  & \tableindent 6.40 & \tableindent 8.40 & \tableindent 6.90 & \\
$\text{ILT}^\dagger$ \cite{michieli2019ilt} & 67.10 & 12.30 & 64.40 &  & 66.30 & 40.60 & 59.90 &  & \tableindent 4.90 & \tableindent 7.80 & \tableindent 5.70 & \\ 
$\text{ILT}$ \cite{michieli2019ilt} & 67.75 & 10.88 & 65.05 & 71.23 & 67.08 & 39.23 & 60.45 & 70.37 & \tableindent 8.75 & \tableindent 7.99 & \tableindent 8.56 & 40.16 \\ 

$\text{MiB}^\dagger$ \cite{cermelli2020modelingthebackground} & 70.20 & 22.10 & 67.80 &    & 75.50 & 49.40 & 69.00 &  & 35.10 & 13.50 & 29.70 & \\
% from us
MiB \cite{cermelli2020modelingthebackground} & 71.43 & 23.59 & 69.15  & 73.28  & \textbf{76.37}  & 49.97  & \textbf{70.08} & \textbf{75.12} & 34.22 & 13.50  & 29.29  & 54.19 \\

\ours & \textbf{75.35} & \textbf{37.35} & \textbf{73.54} & \textbf{75.47} & 75.73 & \textbf{51.71} & \textbf{70.09} & \textbf{75.19} & \textbf{65.12} & \textbf{21.11} & \textbf{54.64} & \textbf{67.21}\\
%\midrule
%\ours vs MiB & +3.92 & +13.76 & +4.39 & +1.19 & -0.64 & +1.84 & +0.01 & +0.07 & +30.90 & +7.61 & +25.35 & +13.02\\
% Algo &   &   &   &   &   &  &  &   &   &   &  &  \\
%\midrule
%Joint model & 77.40 & 78.00 & 77.40 & --- & 79.10 & 72.60 & 77.40 & --- & 79.10 & 72.60 & 77.40 & ---\\
\bottomrule
\end{tabular}
\end{table*}

\begin{table*}[t]
\centering
\caption{Continual Semantic Segmentation results on ADE20k in Mean IoU (\%).}
\vspace*{-0.3cm}
\label{tab:ade_sota}
\begin{tabular}{@{}l|cccc||cccc||cccc@{}}
\toprule
& \multicolumn{4}{c}{\textbf{100-50} (2 tasks)} & \multicolumn{4}{c}{\textbf{50-50} (3 tasks)} & \multicolumn{4}{c}{\textbf{100-10} (6 tasks)}\\
\textbf{Method} & 0-100 & 101-150 & \textit{all} & \textit{avg} & 0-50 & 51-150 & \textit{all} & \textit{avg}  & 0-100 & 101-150 & \textit{all} & \textit{avg} \\
\midrule
ILT \cite{michieli2019ilt} & 18.29  & 14.40  & 17.00  & 29.42  & \tableindent 3.53  & 12.85  & \tableindent 9.70 & 30.12 & \tableindent 0.11  & \tableindent 3.06 & \tableindent 1.09 & 12.56 \\ 
MiB \cite{cermelli2020modelingthebackground} & 40.52  & \textbf{17.17}  & \textbf{32.79}  & \textbf{37.31}  & 45.57  & \textbf{21.01}  & 29.31 & 38.98 & 38.21 & 11.12 & 29.24 & 35.12 \\
\ours  & \textbf{41.87}  & 14.89  & \textbf{32.94}  & \textbf{37.39}  & \textbf{48.83}  & \textbf{20.99}  & \textbf{30.40} & \textbf{39.42} & \textbf{40.48}  & \textbf{13.61} & \textbf{31.59} & \textbf{36.64}\\
% Note that ADE 50-50 was done without adaptive!
% Algo & \tbd  & \tbd  & \tbd  & \tbd  & \tbd  & \tbd & \tbd & \tbd  & \tbd  & \tbd  & \tbd & \tbd \\
%\midrule
%\ours vs MiB & +1.35  & -2.28  & +0.15  & +0.08  & +3.26  & -0.02  & +1.09 & +0.44 & +2.27  & +2.49 & +2.35 & +1.52 \\
%\midrule
% Joint model & 44.30 & 28.20 & 38.90 & --- & 51.10 & 33.25 & 38.90 & ---  & 44.30 & 28.20 & 38.90 & --- \\
\bottomrule
\end{tabular}
\end{table*}

where $u$ represents the uncertainty of pixel $(w,h)$ and $\tau_{c}$ is a class-specific threshold. Thus, we discard all the pixels for which the old model is uncertain ($u \ge \tau_c$) in \autoref{eq:pseudo_bis_uncertain} and decrement the normalization factor $WH$ by one. We use entropy as the uncertainty measurement $u$. Specifically, before learning task $t$, we compute the median entropy for the old model over all pixels of $\mathcal{D}^t$ predicted as $c$ for all the previous classes $c \in C^{1:t-1}$, which provides in thresholds $\tau_c \in C^{1:t-1}$, as proposed in \cite{saporta2020esl}. The cross-entropy loss with pseudo-labeling of the old classes can be written as:

\vspace*{-0.5cm}%sale de fou
\begin{equation}
    \mcL_\text{pseudo}(\Theta^t)=- \frac{\nu}{WH} \sum_{w,h}^{W,H} \sum_{c \in \mathcal{C}^{t}} \tilde S\left(w,h,c\right) \log \hat S^{t}\left(w,h,c\right)\,,
\label{eq:pseudo_loss}
\end{equation}

where $\nu$ is the ratio of accepted old classes pixels over the total number of such pixels. This ponderation allows to adaptively weight the importance of the pseudo-labeling within the total loss. We call \ours (standing for Pseudo-labeling and LOcal Pod) the proposed approach, that uses both Local POD to avoid catastrophic forgetting, and our uncertainty-based pseudo-labeling to address background shift. To sum it up, the total loss in \ours is:

\begin{equation}
    \mcL(\Theta^t) = \underbrace{\strut \mcL_\text{pseudo}(\Theta^t)}_\text{classification} + \lambda\underbrace{\strut \mcL_\text{localPod}(\Theta^t)}_\text{distillation}\,,
\label{eq:complete_loss}
\end{equation}

with $\lambda$ an hyperparameter.

\section{Experiments}

\subsection{Datasets, Protocols, and Baselines}
\label{sec:datasets_protocols}

To ensure fair comparisons with state-of-the-art approaches, we follow the experimental setup of ~\cite{cermelli2020modelingthebackground} for datasets, protocol, metrics, and baseline implementations.

\noindent\textbf{Datasets:\,} we evaluate \ours on 3 segmentation datasets: Pascal-VOC 2012 \cite{everingham2015pascalvoc} (20 classes), ADE20k \cite{zhou2017adedataset} (150 classes) and CityScapes \cite{cordts2016cityscapes} (19 classes from 21 different cities). Full details are in the supplementary materials.

\noindent\textbf{CSS protocols:\,} ~\cite{cermelli2020modelingthebackground} describes two different CSS settings: \textit{Disjoint} and \textit{Overlapped}. In both, only the current classes are labeled vs. a background class $C^t$. However, in the former, images of task $t$ only contain pixels $C^{1:t-1} \cup C^{t}$ (old and current), while, in the latter, pixels can belong to any classes $C^{1:t-1} \cup C^{t} \cup C^{t+1:T}$ (old, current, and future). Thus, the Overlapped setting is the most challenging and realistic, as in a real setting there isn't any oracle method to exclude future classes from the background. Therefore, in our experiments, we focus on Overlapped CSS but more results for Disjoint CSS can be found in the supplementary materials. While the training images are only labeled for the current classes, the testing images are labeled for all seen classes. We evaluate several CSS protocols for each dataset, e.g. on VOC 19-1, 15-5, and 15-1 respectively consists in learning 19 then 1 class ($T=2$ steps), 15 then 5 classes ($2$ steps), and 15 classes followed by five times 1 class ($6$ steps). The last setting is the most challenging due to its higher number of steps. Similarly, on ADE 100-50 means 100 followed by 50 classes ($2$ steps), 100-10 means 100 followed by 5 times 10 classes ($6$ steps), and so on.

\noindent\textbf{Metrics:\,} we compare the different models using traditional mean Intersection over Union (mIoU). Specifically, we compute mIoU after the last step $T$ for the initial classes $C^{1}$, for the incremented classes $C^{2:T}$, and for all classes $C^{1:T}$ (\textit{all}). These metrics respectively reflect the robustness to catastrophic forgetting (the model rigidity), the capacity to learn new classes (plasticity), as well as its overall performance (trade-of between both). We also introduce a novel \textit{avg} metric (short for \textit{average}), which measures the average of mIoU scores measured step after step, integrating performance over the whole continual learning process. 

\noindent\textbf{Baselines:\,}We benchmark our model against the latest state-of-the-arts CSS methods ILT~\cite{michieli2019ilt} and MiB~\cite{cermelli2020modelingthebackground}. We also evaluate general continual models based on weight constraints (EWC \cite{kirkpatrick2017ewc}) and knowledge distillation (LwF-MC \cite{rebuffi2017icarl}). More baselines are available in the supplementary materials. All models, ours included, don't use rehearsal learning \cite{robins1995catastrophicforgetting,rebuffi2017icarl,chaudhry2019tinyepisodicmemories} where a limited quantity of previous tasks data can be rehearsed.
Finally, we also compare with a reference model learned in a traditional semantic segmentation setting (``Joint model'' without continual learning), which may constitute an upper bound for CSS methods.

\noindent\textbf{Implementation Details:\,}As in~\cite{cermelli2020modelingthebackground}, we use a Deeplab-V3~\cite{chen2017deeplabv3} architecture with a ResNet-101~\cite{he2016resnet} backbone pretrained on ImageNet~\cite{deng2009imagenet} for all experiments. Full details are provided in the supplementary materials.

\subsection{Quantitative Evaluation}
\label{sec:quantitative}

First, we compare \ours with state-of-the-art methods.

\begin{figure}
  \includegraphics[width=0.9\linewidth]{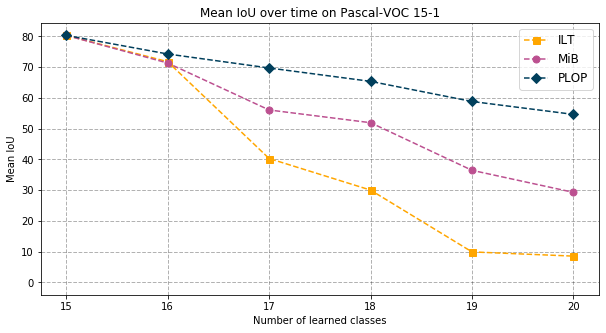} 
     \vspace*{-0.3cm}
    \caption{mIoU evolution on Pascal-VOC 2012 15-1. While MiB's mIoU quickly deteriorates, PLOP's mIoU remains high, due to improved resilience to catastrophic forgetting.}
    \label{fig:plot_voc_15-1}
\end{figure}

\noindent\textbf{Pascal VOC 2012:\,} \autoref{tab:voc_sota} shows quantitative experiments on VOC 19-1, 15-5, and 15-1. \ours outperforms its closest contender, MiB \cite{cermelli2020modelingthebackground} on all evaluated settings by a significant margin. On 19-1, the forgetting of old classes (1-19) is reduced by 4.39 percentage points (\pp) while performance on new classes is greatly improved (+13.76 \pp). On 15-5, our model is on par with our re-implementation of MiB, and surpasses the original paper scores \cite{cermelli2020modelingthebackground} by 1 \pp. On the most challenging 15-1 setting, general continual models (EWC and LwF-MC) and ILT all have very low mIoU. While MiB shows significant improvements, \ours still outperforms it by a wide margin: +86\% on all classes, +90\% on old classes, and +56\% on new classes. Also, the joint model mIoU is $77.40\%$, thus \ours narrows the gap compared to state-of-the-art approaches on every CSS scenario. The average mIoU is also improved by +24\% compared to MiB, indicating that each CSS step benefits from the improvements related to our method. This is echoed by \autoref{fig:plot_voc_15-1}, which shows that while mIoU for both ILT and MiB deteriorates after only a handful of steps, \ours's mIoU remains very high throughout, indicating improved resilience to catastrophic forgetting and background shift.

\noindent\textbf{ADE20k:\,} \autoref{tab:ade_sota} shows experiments on ADE 100-50, 100-10, and 50-50. This dataset is notoriously hard, as the joint model baseline mIoU is only 38.90\%. ILT has poor performance in all three scenarios. \ours shows comparable performance with MiB on the short setting 100-50 (only 2 tasks), improves by 1.09 \textit{p.p} on the medium setting 50-50 (3 tasks), and significantly outperforms MiB with a wider margin of 2.35 \textit{p.p} on the long setting 100-10 (6 tasks). In addition to being better on all settings, PLOP showcased an increased performance gain on longer CSS (e.g. 100-10) scenarios, due to increased robustness to catastrophic forgetting and background shift. To further validate this robustness, we propose harder novel CSS scenarios.

\subsection{New Protocols and Evaluation}

\begin{table}[t]
\centering
\caption{Mean IoU on Pascal-VOC 2012 10-1.}
\vspace*{-0.3cm}
\label{tab:voc_hard}
\begin{tabular}{@{}l|cccc@{}}
\toprule
 & \multicolumn{4}{c}{\textbf{VOC 10-1} (11 tasks)}\\
\textbf{Method} & 0-10 & 11-20 & \textit{all} & \textit{avg}\\
\midrule
ILT \cite{michieli2019ilt} & \tableindent 7.15 & \tableindent 3.67 & \tableindent 5.50 & 25.71\\ 
MiB \cite{cermelli2020modelingthebackground} & 12.25 & 13.09 & 12.65 & 42.67 \\
\ours & \textbf{44.03} & \textbf{15.51} & \textbf{30.45} & \textbf{52.32}\\
\bottomrule
\end{tabular}
\end{table}

\begin{table}[t]
\centering
\caption{Mean IoU on ADE20k 100-5.}
\vspace*{-0.3cm}
\label{tab:ade_hard}
\begin{tabular}{@{}l|cccc@{}}
\toprule
 & \multicolumn{4}{c}{\textbf{ADE 100-5} (11 tasks)}\\
\textbf{Method} & 0-100 & 101-150 & \textit{all} & \textit{avg}\\
\midrule
ILT \cite{michieli2019ilt} & \tableindent 0.08 & \tableindent 1.31 & \tableindent 0.49 & \tableindent 7.83\\ 
MiB \cite{cermelli2020modelingthebackground} & 36.01 & \tableindent 5.66 & 25.96  & 32.69 \\

\ours  & \textbf{39.11} & \tableindent \textbf{7.81} & \textbf{28.75} & \textbf{35.25}\\
\bottomrule
\end{tabular}
\end{table}

\begin{comment}
\begin{table*}[t]
\centering
\caption{Mean IoU on the Pascal-VOC 2012 and ADE20k datasets for harder continual settings.}
\label{tab:voc_ade_hard}
\begin{tabular}{@{}l|cccc||cccc@{}}
\toprule
 & \multicolumn{4}{c}{\textbf{VOC 10-1} (11 tasks)} & \multicolumn{4}{c}{\textbf{ADE 100-5} (11 tasks)}\\
\textbf{Method} & 1-10 & 11-20 & \textit{all} & \textit{avg} & 0-100 & 101-150 & \textit{all} & \textit{avg}\\
\midrule
ILT \cite{michieli2019ilt} & \tableindent 7.15 & \tableindent 3.67 & \tableindent 5.50 & 25.71 & \tableindent 0.08 & \tableindent 1.31 & \tableindent 0.49 & \tableindent 7.83\\ 
MiB \cite{cermelli2020modelingthebackground} & 12.25 & 13.09 & 12.65 & 42.67 & 36.01 & \tableindent 5.66 & 25.96  & 32.69 \\

\ours & \textbf{44.03} & \textbf{15.51} & \textbf{30.45} & \textbf{52.32} & \textbf{39.11} & \tableindent \textbf{7.81} & \textbf{28.75} & \textbf{35.25}\\
% Note that for VOC 10-1 we only do 10 epochs per task
% Algo &  &  &  &  &   &  &  & \\ 
%\midrule
%\ours vs MiB & +31.78 & +2.42 & +17.80 & +9.65 & +3.10 & +2.15 & +2.79 & +2.56\\
%\midrule
%Joint model & 78.76 & 76.69 & 77.40 & --- & 44.30 & 28.20 & 38.90 & ---  \\
\bottomrule
\end{tabular}
\end{table*}
\end{comment}

\noindent\textbf{Longer Continual Learning:\,} We argue that CSS experiments should push towards more steps \cite{wortsman2020supermasks,lomonaco2020ar1,douillard2020podnet,castro2018end_to_end_inc_learn} to quantify the robustness of approaches w.r.t. catastrophic forgetting and background shift. We introduce two novel and much more challenging settings with 11 tasks, almost twice as many as the previous longest setting. We report results for VOC 10-1 in \autoref{tab:voc_hard} (10 classes followed by 10 times 1 class) and ADE 100-5 in \autoref{tab:ade_hard} (100 classes followed by 10 times 5 classes). The second previous State-of-the-Art method, ILT, has a very low mIoU ($<6$ on VOC 10-1 and practically null on ADE 100-5). Furthermore, the gap between \ours and MiB is even wider compared with previous benchmarks (e.g. $\times$3.6 mIoU on VOC for mIoU of base classes 1-10), which confirms the superiority of \ours when dealing with long continual processes.

\begin{figure}
    \includegraphics[width=0.9\linewidth]{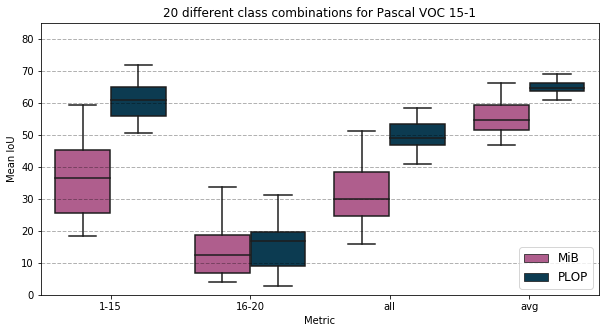}
    \vspace*{-0.3cm}
    \caption{Boxplots of the mIoU of initial classes (1-15), new (16-20), all, and average for 20 random class orderings. PLOP is significantly better and more stable than MiB.}
    \label{fig:order_voc_15-1}
\end{figure}

\noindent\textbf{Stability w.r.t. class ordering:\,}We already showed that existing continual learning methods may be prone to instability. It has already been shown in related contexts \cite{kim2019medic} that class ordering can have a large impact on performance. However, in real-world settings, the optimal class order can never be known beforehand: thus, the performance of an ideal CSS method should be as class order-invariant as possible. In all experiments done so far, this class order has been kept constant, as defined in~\cite{cermelli2020modelingthebackground}. We report results in \autoref{fig:order_voc_15-1} under the form of boxplots obtained by applying 20 random permutations of the class order on VOC 15-1. We report in \autoref{fig:order_voc_15-1} (from left to right) the mIoU for the old, new classes, all classes, and average over CSS steps. In all cases, \ours surpasses MiB in term of avg mIoU. Furthermore, the standard deviation (e.g. 10\% vs 5\% on \textit{all}) is always significantly lower, showing the excellent stability of \ours compared with existing approaches.

\begin{table}[t]
\centering
\caption{Final mIoU for Continual-domain Cityscapes.}
\label{tab:cityscapes_domain}
\begin{tabular}{@{}l|c||c||c@{}}
\toprule
\textbf{Method} & \textbf{11-5} (3 tasks) & \textbf{11-1} (11 tasks) & \textbf{1-1} (21 tasks)\\
\midrule
%EWC \cite{kirkpatrick2017ewc} & 60.62 & 45.24 & 34.22\\
%LWF-MC \cite{rebuffi2017icarl} & 58.90 & 56.92 & 31.24\\
ILT \cite{michieli2019ilt} & 59.14 & 57.75 & 30.11 \\
MiB \cite{cermelli2020modelingthebackground} & 61.51 & 60.02 & 42.15 \\
\ours & \textbf{63.51} & \textbf{62.05} & \textbf{45.24}\\
% Algo &   &   &   &   &   &  &  &   &   &   &  &  \\
%\midrule
%Joint model & 77.40 & 78.00 & 77.40 & --- & 79.10 & 72.60 & 77.40 & --- & 79.10 & 72.60 & 77.40 & ---\\
\bottomrule
\end{tabular}
\end{table}

\begin{comment}
\begin{table}[t]
\centering
\caption{Cityscapes domain\tbd}
\label{tab:cityscapes}
\begin{tabular}{@{}l|cccc||cccc@{}}
\toprule
& \multicolumn{2}{c}{\textbf{11-5} (3 tasks)} & \multicolumn{2}{c}{\textbf{11-1} (1 tasks)}\\
\textbf{Method} & \textit{all} & \textit{avg} & \textit{all} & \textit{avg}\\
\midrule
ILT \cite{michieli2019ilt} & 59.14 & 59.14 & 57.65 & 57.92\\
MiB \cite{cermelli2020modelingthebackground} & 61.51 & 59.10 & 59.17 & 57.86\\
\ours & 63.51 & 61.61 & 61.73 & 61.51\\
%\ours w/o adaptive & \textbf{76.05}  & \textbf{33.34}  & \textbf{74.02} & \textbf{75.71} & 75.98 & \textbf{48.95} & 69.54 & 74.91 & \textbf{69.74} & 12.56 & 56.12 & 66.57\\
%\,\,- LP + UNKD & \textbf{76.67}  & 27.08  & \textbf{74.30} & \textbf{75.85} & \textbf{78.53}  & 48.37  & \textbf{71.35} & \textbf{75.82} & 68.87 & \textbf{21.37} & \textbf{57.56} & \textbf{68.56}\\
\midrule
\ours vs MiB \\
% Algo &   &   &   &   &   &  &  &   &   &   &  &  \\
%\midrule
%Joint model & 77.40 & 78.00 & 77.40 & --- & 79.10 & 72.60 & 77.40 & --- & 79.10 & 72.60 & 77.40 & ---\\
\bottomrule
\end{tabular}
\end{table}
\end{comment}
%\input{tables/cityscapes_class}

\noindent\textbf{Domain Shift:\,} The previous experimental setups mainly assess the capacity of CSS methods to integrate new classes, i.e. to deal with catastrophic forgetting and background shift at a semantic level. However, a domain shift can also happen in CSS scenarios. Thus, we propose a novel benchmark on Cityscapes to quantify robustness to domain shift, in which all 19 classes will be known from the start and, instead of adding new classes, each step brings a novel domain (e.g. a new city), similarly to the NI setting of \cite{lomonaco2017core50} for image classification. \autoref{tab:cityscapes_domain} compares the performance of ILT, MiB, and \ours on CityScapes 11-5, 11-1, and 1-1, making 3, 11 and 21 steps of 11 + 2 times 5 cities, 11 + 10 times 1 city, and 1 + 20 times 1 city respectively. \ours performs better by a significant margin in every such scenario compared with ILT and MiB which, in this setting, is equivalent to a simple cross-entropy plus basic knowledge distillation \cite{hinton2015knowledge_distillation}. Our Local POD, however, retains better domain-related information by modeling long and short-range dependencies at different representation levels.

\subsection{Model Introspection}

We compare several distillation and classification losses on VOC 15-1 to stress the importance of the components of \ours and report results in \autoref{tab:ablation_distill_classif}. All comparisons are evaluated on a val set made with 20\% of the train set, therefore results are slightly different from the main experiments.

\noindent\textbf{Distillation comparisons:\,}\autoref{tab:ablation_distillation} compares different distillation losses when combined with our pseudo-labeling loss. As such, UNKD introduced in \cite{cermelli2020modelingthebackground} performs better than the Knowledge Distillation (KD) of \cite{hinton2015knowledge_distillation}, but not at every step (as indicated by the \textit{avg.} value), which indicates instability during the training process. POD, proposed in \cite{douillard2020podnet}, improves the results on the old classes, but not on the new classes (16-20). In fact, due to too much plasticity, POD model likely overfits and predicts nothing but the new classes, hence a lower mIoU.  Finally, Local POD leads to superior performance (+20 \pp) w.r.t. all metrics, due to its integration of both long and short-range dependencies. This final row represents our full \ours strategy.

\noindent\textbf{Classification comparisons:\,}\autoref{tab:ablation_classif} compares different classification losses when combined with our Local POD distillation loss. Cross-Entropy (CE) variants perform poorly, especially on new classes. UNCE, introduced in \cite{cermelli2020modelingthebackground}, improves by merging the background with old classes, however, it still struggles to correctly model the new classes, whereas our pseudo-labeling propagates more finely information of the old classes, while learning to predict the new ones, dramatically enhancing the performance in both cases. This penultimate row represents our full \ours strategy.
Also notice that the performance for pseudo-labeling is very close to \textit{Pseudo-Oracle} (where the incorrect pseudo-labels are removed), which may constitute a performance ceiling of our uncertainty measure. A comparison between these two results illustrates the relevance of our entropy-based uncertainty estimate.

\begin{table}
\centering
\caption{Comparison studies on Pascal-VOC 2012 15-1 on a validation subset of 20\% of the training set.}
\vspace*{-0.3cm}
\label{tab:ablation_distill_classif}
\begin{subtable}{0.5\textwidth}
    \centering
    \caption{Pseudo loss (\autoref{eq:pseudo_loss}) with different distillation losses.}
    \vspace*{-0.2cm}
    \label{tab:ablation_distillation}
    \begin{tabular}{@{}l|cccc@{}}
    \toprule
    Distillation loss & 0-15 & 16-20 & \textit{all} & \textit{avg}\\
    \midrule
    %ILT \cite{michieli2019ilt}'s distill & 19.91 & \tableindent 5.49 & 16.48 & 49.43\\
    %CSC \cite{park2020csc} & 25.49 & \tableindent 4.72 & 20.48 & 44.97\\
    Knowledge Distillation & 29.72 & \tableindent 4.42 & 23.69 & 49.18\\
    UNKD & 34.85 & \tableindent 5.26 & 27.80 & 46.39\\
    POD & 43.94 & \tableindent 4.82 & 34.62 & 53.35\\
    Local POD (\autoref{eq:local_pod_loss}) & \textbf{63.06} & \textbf{17.92} & \textbf{52.31} & \textbf{65.71}\\
    \bottomrule
    \end{tabular}
\end{subtable}
\hfill
\begin{subtable}{0.5\textwidth}
    \centering
    \caption{Local POD loss (\autoref{eq:local_pod_loss}) with different classification losses.}
    \vspace*{-0.2cm}
    \label{tab:ablation_classif}
    \begin{tabular}{@{}l|cccc@{}}
    \toprule
    Classification loss & 0-15 & 16-20 & \textit{all} & \textit{avg}\\
    \midrule
    CE only on new & 12.95 & \tableindent 2.54 & 10.47 & 47.02 \\
    CE & 33.80 & \tableindent 4.67 & 26.87 & 50.79 \\
    UNCE & 48.46 & \tableindent 4.82 & 38.62 & 53.19 \\
    Pseudo (\autoref{eq:pseudo_loss}) & \textbf{63.06} & \textbf{17.92} & \textbf{52.31} & \textbf{65.71}\\
    \midrule
    \textit{\small{Pseudo-Oracle}} & \textit{63.69} & \textit{23.35} & \textit{54.09} & \textit{66.05}\\
    %\textit{\small{Pseudo + corrected}} & \textit{66.88} & \textit{16.88} & \textit{54.98} & \textit{66.50}\\
    %\textit{\small{CE + all labels}}  & \textit{71.45} & \textit{10.78} & \textit{57.00} & \textit{67.04}\\
    \bottomrule
    \end{tabular}
\end{subtable}
\end{table}

\begin{figure*}
  \centering
  \includegraphics[width=\linewidth]{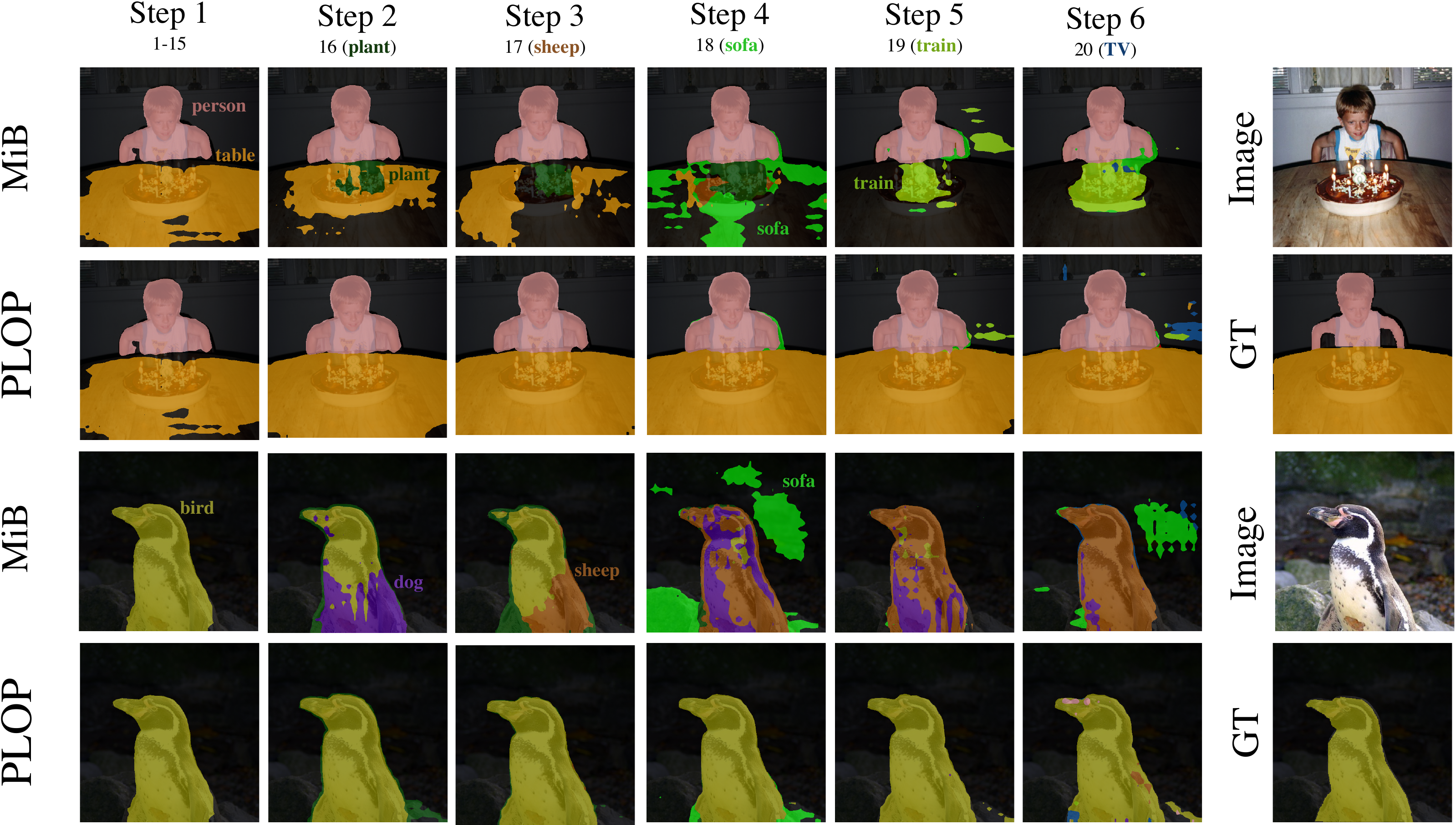}
    \caption{Visualization of MiB and PLOP predictions across time in VOC 15-1 for two test images. MiB quickly forgets the initial 15 classes (row 1: \texttt{person} and \texttt{table}, row 3: \texttt{bird}) in favor of new classes (\texttt{plant}, \texttt{sheep}, \texttt{sofa}, \texttt{train}) and is biased towards new classes. PLOP, however, barely suffers from catastrophic forgetting (rows 2+4).}
    \label{fig:visualization}
\end{figure*}

\begin{figure}
  \includegraphics[width=\linewidth]{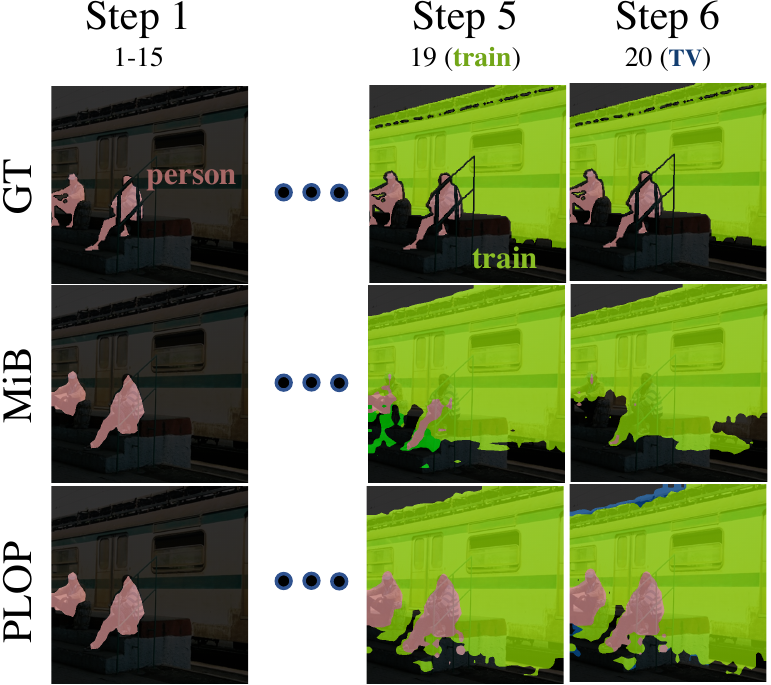}
    \caption{Visualization of MiB and PLOP predictions across time in VOC 15-1 on a test set image. At steps 1-4 only class \texttt{person} has been seen. At step 5, the class \texttt{train} is introduced, causing dramatic background shift. While MiB overfits on the new class and forget the old class, PLOP is able to predict both classes correctly.}
    \label{fig:visualization_gt_shift}
\end{figure}

\noindent\textbf{Vizualisation:\,} \autoref{fig:visualization} shows the predictions for both MiB and PLOP on VOC 15-1 across time. At first, both models output equivalent predictions. However, MiB quickly forgets the previous classes and becomes biased towards new classes. On the other hand, PLOP predictions are much more stable on old classes while learning new classes, thanks to Local POD alleviating catastrophic forgetting by spatially constraining representations, and pseudo-labeling dealing with background shift. \autoref{fig:visualization_gt_shift} more closely highlights this phenomenon: at first, the ground-truth only contains the class \texttt{person}. At step $5$, the class \texttt{train} is introduced. As a result, MiB overfits on \texttt{train} and forgets \texttt{person}. PLOP, instead, manages to avoid forgetting \texttt{person} and predicts decent segmentation for both classes.
\section{Conclusion}

In this paper, we paved the way for future research on Continual Semantic Segmentation, which is an emerging domain in computer vision. We highlighted two main challenges in Continual Semantic Segmentation (CSS), namely catastrophic forgetting and background shift. To deal with the former, we proposed Local POD, a multi-scale pooling distillation scheme that allows preserving long and short-range spatial relationships between pixels, leading to a suitable trade-off between rigidity and plasticity for CSS and, ultimately, alleviating catastrophic forgetting. The proposed method is general enough to be used in other related distillation settings, where preserving spatial information is a concern. In addition, we introduced a new strategy to address the background shift based on an efficient pseudo-labeling method. We validate our \ours framework, on several existing CSS scenarios involving multiple datasets. In addition, we propose novel experimental scenarios to assess the performance of future CSS approaches in terms of long term learning capacity and stability. We showed that \ours performs significantly better than all existing baselines in every such CSS benchmark.

{\footnotesize \noindent\textbf{Acknowledgments}: This  work  was  granted  access  to  the  HPC  resources  of IDRIS under the allocation AD011011706 made by GENCI.}
%\clearpage\newpage

\appendix
\section{Appendix}

\subsection{Further Work}

In our CSS setting, pixels of task $T$ can belong to old $C^{1:t-1}$, current $C^t$, and future classes $C^{t+1:T}$. In this paper we cover how to better handle old and current classes. Further works should investigate how to exploit the already present future information with Zeroshot \cite{lampert2009zeroshot,kumar2018synthesized_zeroshot} as already done in semantic segmentation \cite{kato2019zeroshotsegmentation,bucher2019zeroshotsegmentation} and explored for continual classification \cite{wang2020bookworm,douillard2020ghost}.

\subsection{Algorithm view of Local POD}

In \autoref{algo:local_pod}, we summarize the algorithm for the proposed Local POD. The algorithm consists in three functions. First, \texttt{Distillation}, loops over all $L$ layers onto which we apply Local POD. Second, \texttt{LocalPOD}, computes the L2 distance (L.26) between POD embeddings of the current (L.19) and old (L.20) models. It loops over $S$ different scales (L.14) and $\Phi$ computes the POD embedding given two features maps subsets (L.19-20) as defined in Eq. 1. $\|=$ denotes an in-place concatenation.

\begin{algorithm}
\caption{Local POD algorithm}
\label{algo:local_pod}
\begin{algorithmic}[1]
  \Function{Distillation}{$f^t$, $f^{t-1}$, $x$, $S$}
    \State $loss \gets 0$
    \For{\texttt{$l \gets 0$; $l < L$; $l{+}{+}$}}
      \State $\vh^t_l \gets f^t_l(\vx)$
      \State $\vh^{t-1}_l \gets f^{t-1}_l(\vx)$
      
      \State $loss \gets loss + \operatorname{LocalPOD}(\vh^t_l, \vh^{t-1}_l, S)$
    \EndFor
    \State \Return $\frac{loss}{L}$
  \EndFunction
  \\
  \Function{LocalPOD}{$\vh^t$, $\vh^{t-1}, S$} 
    \State $\mathbf{P}^t \gets [\,]$
    \State $\mathbf{P}^{t-1} \gets [\,]$
    
    \For{\texttt{$s \gets 0$; $s < S$; $s{+}{+}$}} \Comment{Eq. 3}
        \State $w \gets \nicefrac{W}{2^s}$
        \State $h \gets \nicefrac{H}{2^s}$

        \For{\texttt{$i \gets 0$; $i < W - w$; $i{+}=w$}}
            \For{\texttt{$j \gets 0$; $j < H - h$; $j{+}=h$}}
                \State $\vp^t \gets \operatorname{\Phi}(\vh^t\texttt{[i:i+w, j:j+h]})$ \Comment{Eq. 1}
                \State $\vp^{t-1} \gets \operatorname{\Phi}(\vh^{t-1}\texttt{[i:i+w, j:j+h]})$ 

                \State $\mathbf{P}^t \|= \vp^t$
                \State $\mathbf{P}^{t-1} \|= \vp^{t-1}$
            \EndFor
        \EndFor
    \EndFor
    \State \Return $\left\Vert \mathbf{P}^t - \mathbf{P}^{t-1}\right\Vert^2$ \Comment{Eq. 5}
  \EndFunction
\end{algorithmic}
\end{algorithm}

\subsection{Reproducibility}

\noindent\textbf{Datasets:\,} We evaluate our model on three datasets Pascal-VOC~\cite{everingham2015pascalvoc}, ADE20k~\cite{zhou2017adedataset}, and Cityscapes~\cite{cordts2016cityscapes}. VOC contains 20 classes, 10,582 training images, and 1,449 testing images. ADE20k has 150 classes, 20,210 training images, and 2,000 testing images. Cityscapes contains 2975 and 500 images for train and test, respectively. Those images represent 19 classes and were taken from 21 different cities. All ablations and hyperparameters tuning were done on a validation subset of the training set made of 20\% of the images. 
For all datasets, we resize the images to $512 \times 512$, with a center crop. An additional random horizontal flip augmentation is applied at training time.

\noindent\textbf{Implementation details:\,} For all experiments, we use a Deeplab-V3~\cite{chen2017deeplabv3} architecture with a ResNet-101~\cite{he2016resnet} backbone pretrained on ImageNet~\cite{deng2009imagenet}, as in~\cite{cermelli2020modelingthebackground}. For all datasets, we set a maximum threshold for the uncertainty measure of Eq. 7 to $\tau=1e-3$. We train our model for 30 and 60 epochs per CSS step on Pascal VOC and ADE, respectively, with an initial learning rate of $1e-2$ for the first CSS step, and $1e-3$ for all the following ones. We reduce the learning rate exponentially with a decay rate of $9e-1$. We use SGD optimizer with $9e-1$ Nesterov momentum. The Local POD factor $\lambda$ is set to $1e-2$ and $5e-4$ for intermediate feature maps and logits, respectively. Moreover, we multiply this factor by the adaptive weighting $\sqrt{\nicefrac{|C^{1:t}|}{|C^{t}|}}$ introduced by \cite{hou2019ucir} that increases the strength of the distillation the further we are into the continual process. For all feature maps, Local POD is applied before ReLU, with squared pixel values, as in \cite{zagoruyko2016distillation_attention,douillard2020podnet}. We use 3 scales for Local POD: $1$, $\nicefrac{1}{2}$, and $\nicefrac{1}{4}$, as adding more scales experimentally brought diminishing returns. We use a batch size of 24 distributed on two GPUs. Contrary to many continual models, we don't have access to any task id in inference, therefore our setting/strategy has to predict a class among the set of all seen classes ---a realist setting.

% \subsection{Classes ordering details}

\noindent\textbf{Classes ordering details:\,} For all quantitative experiments on Pascal-VOC 2012 and ADE20k, the same class ordering was used across all evaluated models. For Pascal-VOC 2012 it corresponds to \lstinline![1, 2, ..., 20]! and ADE20k to \lstinline![1, 2, ..., 150]! as defined in \cite{cermelli2020modelingthebackground}. For continual-domain cityscapes, the order of the domains/cities is the following: \texttt{aachen}, \texttt{bremen}, \texttt{darmstadt}, \texttt{erfurt}, \texttt{hanover}, \texttt{krefeld}, \texttt{strasbourg}, \texttt{tubingen}, \texttt{weimar}, \texttt{bochum}, \texttt{cologne}, \texttt{dusseldorf}, \texttt{hamburg}, \texttt{jena}, \texttt{monchengladbach}, \texttt{stuttgart}, \texttt{ulm}, \texttt{zurich}, \texttt{frankfurt}, \texttt{lindau}, and \texttt{munster}.

In the main paper we showcased a boxplot featuring 20 different class orders for Pascal-VOC 2012 15-1. For the sake of reproducibility, we provide details on these orders:

\begin{adjustbox}{width=\columnwidth,center}
\begin{lstlisting}
[1, 2, 3, 4, 5, 6, 7, 8, 9, 10, 11, 12, 13, 14, 15, 16, 17, 18, 19, 20]
[12, 9, 20, 7, 15, 8, 14, 16, 5, 19, 4, 1, 13, 2, 11, 17, 3, 6, 18, 5]
[9, 12, 13, 18, 2, 11, 15, 17, 10, 8, 4, 5, 20, 16, 6, 14, 19, 1, 7, 3]
[13, 19, 15, 17, 9, 8, 5, 20, 4, 3, 10, 11, 18, 16, 7, 12, 14, 6, 1, 2]
[15, 3, 2, 12, 14, 18, 20, 16, 11, 1, 19, 8, 10, 7, 17, 6, 5, 13, 9, 4]
[7, 13, 5, 11, 9, 2, 15, 12, 14, 3, 20, 1, 16, 4, 18, 8, 6, 10, 19, 17]
[12, 9, 19, 6, 4, 10, 5, 18, 14, 15, 16, 3, 8, 7, 11, 13, 2, 20, 17, 1]
[13, 10, 15, 8, 7, 19, 4, 3, 16, 12, 14, 11, 5, 20, 6, 2, 18, 9, 17, 1]
[3, 14, 13, 1, 2, 11, 15, 17, 7, 8, 4, 5, 9, 16, 19, 12, 6, 18, 10, 20]
[1, 14, 9, 5, 2, 15, 8, 20, 6, 16, 18, 7, 11, 10, 19, 3, 4, 17, 12, 13]
[16, 13, 1, 11, 12, 18, 6, 14, 5, 3, 7, 9, 20, 19, 15, 4, 2, 10, 8, 17]
[10, 7, 6, 19, 16, 8, 17, 1, 14, 4, 9, 3, 15, 11, 12, 2, 18, 20, 13, 5]
[7, 5, 3, 9, 13, 12, 14, 19, 10, 2, 1, 4, 16, 8, 17, 15, 18, 6, 11, 20]
[18, 4, 14, 17, 12, 10, 7, 3, 9, 1, 8, 15, 6, 13, 2, 5, 11, 20, 16, 19]
[5, 4, 13, 18, 14, 10, 19, 15, 7, 9, 3, 2, 8, 16, 20, 1, 12, 11, 6, 17]
[9, 12, 13, 18, 7, 1, 15, 17, 10, 8, 4, 5, 20, 16, 6, 14, 19, 11, 2, 3]
[3, 14, 13, 18, 2, 11, 15, 17, 10, 8, 4, 5, 20, 16, 6, 12, 19, 1, 7, 9]
[7, 5, 9, 1, 15, 18, 14, 3, 20, 10, 4, 19, 11, 17, 16, 12, 8, 6, 2, 13]
[3, 14, 6, 1, 2, 11, 12, 17, 7, 20, 4, 5, 9, 16, 19, 15, 13, 18, 10, 8]
[1, 2, 12, 14, 6, 19, 18, 17, 5, 20, 8, 4, 9, 16, 10, 3, 15, 13, 11, 7]
\end{lstlisting}
\end{adjustbox}

In the 15-1 setting, we first learn the first fifteen classes, then increment the five remaining classes one by one. Note that the special class \texttt{background} (0) is always learned during the first task.

\noindent\textbf{Hardware and Code:\,} For each experiment, we used two Titan Xp GPUs with 12 Go of VRAM each. The initial step $t=1$ for each setting is common to all models, therefore we re-use the weights trained on this step. All models took less than 2 hours to train on Pascal-VOC 2012 15-1, and less than 16 hours on ADE20k 100-10. We distributed the batch size equally on both GPUs. All models are implemented in PyTorch~\cite{paszke2017pytorch} and runned with half-precision for efficiency reasons with Nvdia's APEX library (\href{https://github.com/NVIDIA/apex}{https://github.com/NVIDIA/apex}) using O1 optimization level. Our code base is based on \cite{cermelli2020modelingthebackground}'s code (\href{https://github.com/fcdl94/MiB}{https://github.com/fcdl94/MiB}) that we modified to implement our strategy. It is available at \href{https://github.com/arthurdouillard/CVPR2021_PLOP}{https://github.com/arthurdouillard/CVPR2021\_PLOP}.

\subsection{Additional Experiments}

\noindent\textbf{Model ablation:\,} \autoref{tab:ablation} shows the construction of our model component by component on Pascal-VOC 2012 in 15-5 and 15-1. For this experiment, we train our model on 80\% of the training set and evaluate on the validation set made of the remaining 20\%. We report the mIoU at the final task (``\textit{all}'') and the average of the mIoU after each task (``\textit{avg}'').
We start with a crude baseline made of solely cross-entropy (CE). Pseudo-labeling by itself increases by a large margin performance (eg. 3.99 to 19.74 for 15-1). Applying Local POD reduces drastically the forgetting leading to a massive gain of performance (eg. 19.74 to 50.41 for 15-1). Finally our adaptive factor $\nu$ based on the ratio of accepted pseudo-labels over the number of background pixels further increases our overall results (eg. 50.41 to 52.31 for 15-1). The interest of $\nu$ arises when PLOP faces hard images where few pseudo-labels will be created due to an overall high uncertainty. In such a case, current classes will be over-represented, which can in turn lead to strong bias towards new classes (\textit{i.e.} the model will have a tendency to predict one of the new classes for every pixel). The $\nu$ factor therefore decreases the overall classification loss on such images, and empirical results confirm its effectiveness.

\begin{table}[t]
\centering
\caption{Ablations of \ours on the Pascal-VOC 2012 dataset in 15-5 and 15-1. Scores are measured on a validation subset made of 20\% of the training set.}
\label{tab:ablation}
\begin{tabular}{@{}l||cc|cc@{}}
\toprule
 &  \multicolumn{2}{c}{\textbf{15-5} (2 tasks)} & \multicolumn{2}{c}{\textbf{15-1} (6 tasks)}\\
\cmidrule{2-5}
Model & \textit{all} & \textit{avg} & \textit{all} & \textit{avg}\\
\midrule
CE  & 13.85 & 46.91 & \tableindent 3.99 & 19.37 \\
Pseudo & 66.19 & 73.07 & 19.74 & 44.48 \\
Pseudo + Local POD & 70.29 & 75.13 & 50.41 & 64.95 \\
$\nu$Pseudo + Local POD & \textbf{71.43} & \textbf{75.70} & \textbf{52.31} & \textbf{65.71}\\
\bottomrule
\end{tabular}
\end{table}

\noindent\textbf{Pascal-VOC 2012 Disjoint:\,} In the main paper, we reported results on Pascal-VOC 2012 Overlap. For reasons mentioned previously, Overlap is a more realist setting than Disjoint. Nevertheless, for the sake of comparison, we also provide results in \autoref{tab:voc_disjoint_sota} in the Disjoint setting.
While PLOP has similar performance to MiB in 15-5 (the differences are not significant), it significantly outperforms previous state-of-the-art methods in both 19-1 and 15-1.

\begin{table*}[t]
\centering
\caption{Mean IoU on the Pascal-VOC 2012 dataset for different incremental class learning scenarios, all in Disjoint. $\dagger$ denotes results from Cermelli et al.\cite{cermelli2020modelingthebackground}.}
\label{tab:voc_disjoint_sota}
\begin{tabular}{@{}l|cccc||cccc||cccc@{}}
\toprule
& \multicolumn{4}{c}{\textbf{19-1} (2 tasks)} & \multicolumn{4}{c}{\textbf{15-5} (2 tasks)} & \multicolumn{4}{c}{\textbf{15-1} (6 tasks)}\\
\textbf{Method} & 0-19 & 20 & \textit{all} & \textit{avg} & 0-15 & 16-20 & \textit{all} & \textit{avg} & 0-15 & 16-20 & \textit{all} & \textit{avg}\\
\midrule
% from paper MiB
$\text{Fine Tuning}^\dagger$ & \tableindent 5.80 & 12.30 & \tableindent 6.20 & & \tableindent 1.10 & 33.60 & \tableindent 9.20 & & \tableindent 0.20 & \tableindent 1.80 & \tableindent 0.60 \\
$\text{PI}^\dagger$ \cite{zenke2017synaptic_intelligence} & \tableindent 5.40 & 14.10 & \tableindent 5.90 & & \tableindent 1.30 & 34.10 & \tableindent 9.50 & & \tableindent 0.00 & \tableindent 1.80 & \tableindent 0.40 & \\
$\text{EWC}^\dagger$ \cite{kirkpatrick2017ewc} & 23.20 & 16.00 & 22.90 & & 26.70 & 37.70 & 29.40 & & \tableindent 0.30 & \tableindent 4.30 & \tableindent 1.30 & \\
$\text{RW}^\dagger$ \cite{chaudhry2018riemannien_walk} & 19.40 & 15.70 & 19.20 & & 17.90 & 36.90 & 22.70 & & \tableindent 0.20 & \tableindent 5.40 & \tableindent 1.50 & \\
$\text{LwF}^\dagger$ \cite{li2018lwf} & 53.00 & \tableindent 9.10 & 50.80 & & 58.40 & 37.40 & 53.10 & & \tableindent 0.80 & \tableindent 3.60 & \tableindent 1.50 & \\
$\text{LwF-MC}^\dagger$ \cite{rebuffi2017icarl} & 63.00 & 13.20 & 60.50 & & 67.20 & 41.20 & 60.70 & & \tableindent 4.50 & \tableindent 7.00 & \tableindent 5.20 & \\
$\text{ILT}^\dagger$ \cite{michieli2019ilt} & 69.10 & 16.40 & 66.40 & & 63.20 & 39.50 & 57.30 & & \tableindent 3.70 & \tableindent 5.70 & \tableindent 4.20 &  \\ 
%$\text{ILT}$ \cite{michieli2019ilt} & 71.17 & 16.12 & 68.55 & 73.59 & 64.92 & 38.75 & 58.69 & 69.15 & 10.17 & \tableindent 8.31 & \tableindent 9.73 & 41.05\\ 

$\text{MiB}^\dagger$ \cite{cermelli2020modelingthebackground} & 69.60 & 25.60 & 67.40 & & \textbf{71.80} & \textbf{43.30} & \textbf{64.70} & & 46.20 & 12.90 & 37.90 & \\
% from us
%MiB \cite{cermelli2020modelingthebackground} & 71.22 & 20.95 & 68.83 & 73.60 & \textbf{73.68} & \textbf{45.32} & \textbf{66.93} & \textbf{72.99} & 40.99 & 12.68 & 34.25 & 58.09\\

\ours & \textbf{75.37} & \textbf{38.89} & \textbf{73.64} & 75.71 & \textbf{71.00} & \textbf{42.82} & \textbf{64.29} & 72.05 & \textbf{57.86} & \textbf{13.67} & \textbf{46.48} & 62.67\\
%\midrule
%\ours vs MiB &   &   &   &   &   &  &  &   &   &   &  &\\
% Algo &   &   &   &   &   &  &  &   &   &   &  &  \\
%\midrule
%Joint model & 77.40 & 78.00 & 77.40 & --- & 79.10 & 72.60 & 77.40 & --- & 79.10 & 72.60 & 77.40 & ---\\
\bottomrule
\end{tabular}
\end{table*}

\noindent\textbf{Pascal-VOC 2012 Overlap with more baselines:\,} In \autoref{tab:voc_sota_full}, we report results on Pascal-VOC 2012 Overlap with more baselines. In addition to the models presented in the main paper, we add a naive Fine Tuning, two continual models based on weights constraints (PI \cite{zenke2017synaptic_intelligence} and RW \cite{chaudhry2018riemannien_walk}), and one continual model based on knowledge distillation (LwF \cite{li2018lwf}). PLOP surpasses these methods in all CSS scenarios. 

\begin{table*}[t]
\centering
\caption{Mean IoU on the Pascal-VOC 2012 dataset for different incremental class learning scenarios, all in Overlap. $\dagger$ denotes results from Cermelli et al.~\cite{cermelli2020modelingthebackground}, all other results are from us.}
\label{tab:voc_sota_full}
\begin{tabular}{@{}l|cccc||cccc||cccc@{}}
\toprule
& \multicolumn{4}{c}{\textbf{19-1} (2 tasks)} & \multicolumn{4}{c}{\textbf{15-5} (2 tasks)} & \multicolumn{4}{c}{\textbf{15-1} (6 tasks)}\\
\textbf{Method} & 0-19 & 20 & \textit{all} & \textit{avg} & 0-15 & 16-20 & \textit{all} & \textit{avg} & 0-15 & 16-20 & \textit{all} & \textit{avg}\\
\midrule
% from paper MiB
$\text{Fine Tuning}^\dagger$ & \tableindent 6.80 & 12.90 & \tableindent 7.10 &  & \tableindent 2.10 & 33.10 & \tableindent 9.80 &  & \tableindent 0.20 & \tableindent 1.80 & \tableindent 0.60 & \\
$\text{PI}^\dagger$ \cite{zenke2017synaptic_intelligence} & \tableindent 7.50 & 14.00 & \tableindent 7.80 &  & \tableindent 1.60 & 33.30 & \tableindent 9.50 &  & \tableindent 0.00 & \tableindent 1.80 & \tableindent 0.50 & \\
$\text{EWC}^\dagger$ \cite{kirkpatrick2017ewc} & 26.90 & 14.00 & 26.30 &  & 24.30 & 35.50 & 27.10 &  & \tableindent 0.30 & \tableindent 4.30 & \tableindent 1.30 &  \\
$\text{RW}^\dagger$ \cite{chaudhry2018riemannien_walk} & 23.30 & 14.20 & 22.90 &  & 16.60 & 34.90 & 21.20 &  & \tableindent 0.00 & \tableindent 5.20 & \tableindent 1.30 & \\
$\text{LwF}^\dagger$ \cite{li2018lwf} & 51.20 & \tableindent 8.50 & 49.10 &  & 58.90 & 36.60 & 53.30 &  & \tableindent 1.00 & \tableindent 3.90 & \tableindent 1.80 & \\
$\text{LwF-MC}^\dagger$ \cite{rebuffi2017icarl} & 64.40 & 13.30 & 61.90 &  & 58.10 & 35.00 & 52.30 &  & \tableindent 6.40 & \tableindent 8.40 & \tableindent 6.90 & \\
$\text{ILT}^\dagger$ \cite{michieli2019ilt} & 67.10 & 12.30 & 64.40 &  & 66.30 & 40.60 & 59.90 &  & \tableindent 4.90 & \tableindent 7.80 & \tableindent 5.70 & \\ 
$\text{ILT}$ \cite{michieli2019ilt} & 67.75 & 10.88 & 65.05 & 71.23 & 67.08 & 39.23 & 60.45 & 70.37 & \tableindent 8.75 & \tableindent 7.99 & \tableindent 8.56 & 40.16 \\ 

$\text{MiB}^\dagger$ \cite{cermelli2020modelingthebackground} & 70.20 & 22.10 & 67.80 &    & 75.50 & 49.40 & 69.00 &  & 35.10 & 13.50 & 29.70 & \\
% from us
MiB \cite{cermelli2020modelingthebackground} & 71.43 & 23.59 & 69.15  & 73.28  & \textbf{76.37}  & 49.97  & \textbf{70.08} & \textbf{75.12} & 34.22 & 13.50  & 29.29  & 54.19 \\

\ours & \textbf{75.35} & \textbf{37.35} & \textbf{73.54} & \textbf{75.47} & 75.73 & \textbf{51.71} & \textbf{70.09} & \textbf{75.19} & \textbf{65.12} & \textbf{21.11} & \textbf{54.64} & \textbf{67.21}\\
%\midrule
%\ours vs MiB & +3.92 & +13.76 & +4.39 & +1.19 & -0.64 & +1.84 & +0.01 & +0.07 & +30.90 & +7.61 & +25.35 & +13.02\\
% Algo &   &   &   &   &   &  &  &   &   &   &  &  \\
%\midrule
%Joint model & 77.40 & 78.00 & 77.40 & --- & 79.10 & 72.60 & 77.40 & --- & 79.10 & 72.60 & 77.40 & ---\\
\bottomrule
\end{tabular}
\end{table*}

\newpage

{\small
\bibliographystyle{ieee_fullname}
\bibliography{egbib}

\begin{thebibliography}{10}\itemsep=-1pt

\bibitem{aljundi2018MemoryAwareSynapses}
Rahaf Aljundi, Francesca Babiloni, Mohamed Elhoseiny, Marcus Rohrbach, and
  Tinne Tuytelaars.
\newblock Memory aware synapses: Learning what (not) to forget.
\newblock In {\em Proceedings of the IEEE European Conference on Computer
  Vision (ECCV)}, 2018.

\bibitem{badrinarayanan2017segnet}
V. Badrinarayanan, A. Kendall, and R. Cipolla.
\newblock Segnet: A deep convolutional encoder-decoder architecture for image
  segmentation.
\newblock {\em IEEE Transactions on Pattern Analysis and Machine Intelligence
  (TPAMI)}, 2017.

\bibitem{belouadah2019il2m}
Eden Belouadah and Adrian Popescu.
\newblock Il2m: Class incremental learning with dual memory.
\newblock In {\em Proceedings of the IEEE International Conference on Computer
  Vision (ICCV)}, 2019.

\bibitem{belouadah2020scail}
Eden Belouadah and Adrian Popescu.
\newblock Scail: Classifier weights scaling for class incremental learning.
\newblock In {\em Proceedings of the IEEE Winter Conference on Application of
  Computer Vision (WACV)}, 2020.

\bibitem{bucher2019zeroshotsegmentation}
Maxime Bucher, Tuan-Hung Vu, Matthieu Cord, and Patrick Pérez.
\newblock Zero-shot semantic segmentation.
\newblock In {\em Advances in Neural Information Processing Systems (NeurIPS)},
  2019.

\bibitem{caesar2018cocoostuff}
Holger Caesar, Jasper R.~R. Uijlings, and Vittorio Ferrari.
\newblock Coco-stuff: Thing and stuff classes in context.
\newblock In {\em Proceedings of the IEEE Conference on Computer Vision and
  Pattern Recognition (CVPR)}, 2018.

\bibitem{castro2018end_to_end_inc_learn}
Francisco~M. Castro, Manuel~J Mar{\'i}n-Jim{\'e}nez, Nicol{\'a}s Guil, Cordelia
  Schmid, and Karteek Alahari.
\newblock End-to-end incremental learning.
\newblock In {\em Proceedings of the IEEE European Conference on Computer
  Vision (ECCV)}, 2018.

\bibitem{cermelli2020modelingthebackground}
Fabio Cermelli, Massimiliano Mancini, Samuel Rota~Bulò, Elisa Ricci, and
  Barbara Caputo.
\newblock Modeling the background for incremental learning in semantic
  segmentation.
\newblock In {\em Proceedings of the IEEE Conference on Computer Vision and
  Pattern Recognition (CVPR)}, 2020.

\bibitem{chaudhry2018riemannien_walk}
Arslan Chaudhry, Puneet Dokania, Thalaiyasingam Ajanthan, and Philip
  H.~S.~Torr.
\newblock Riemannian walk for incremental learning: Understanding forgetting
  and intransigence.
\newblock {\em Proceedings of the IEEE European Conference on Computer Vision
  (ECCV)}, 2018.

\bibitem{chaudhry2019AGEM}
Arslan Chaudhry, Marc’Aurelio Ranzato, Marcus Rohrbach, and Mohamed
  Elhoseiny.
\newblock Efficient lifelong learning with a-gem.
\newblock In {\em Proceedings of the International Conference on Learning
  Representations (ICLR)}, 2019.

\bibitem{chaudhry2019tinyepisodicmemories}
Arslan Chaudhry, Marcus Rohrbach, Mohamed Elhoseiny, Thalaiyasingam Ajanthan,
  Puneet~K. Dokania, Philip~H.S. Torr, and Marc'Aurelio Ranzato.
\newblock On tiny episodic memories in continual learning.
\newblock In {\em International Conference on Machine Learning (ICML)
  Workshop}, 2019.

\bibitem{chen2018ZPSA}
Liang{-}Chieh Chen, Yukun Zhu, George Papandreou, Florian Schroff, and Hartwig
  Adam.
\newblock Encoder-decoder with atrous separable convolution for semantic image
  segmentation.
\newblock In {\em Proceedings of the IEEE European Conference on Computer
  Vision (ECCV)}, 2018.

\bibitem{chen2018deeplab}
Liang-Chieh Chen, George Papandreou, Iasonas Kokkinos, Kevin Murphy, and Alan
  L.~Yuille.
\newblock Deeplab: Semantic image segmentation with deep convolutional nets,
  atrous convolution, and fully connected crfs.
\newblock In {\em IEEE Transactions on Pattern Analysis and Machine
  Intelligence (TPAMI)}, 2018.

\bibitem{chen2017deeplabv3}
Liang-Chieh Chen, George Papandreou, Florian Schroff, and Hartwig Adam.
\newblock Rethinking atrous convolution for semantic image segmentation.
\newblock In {\em arXiv preprint library}, 2017.

\bibitem{cordts2016cityscapes}
M. Cordts, M. Omran, S. Ramos, T. Reheld, M. Enzweiler, R. Benenson, U. Franke,
  S. Roth, and B. Schiele.
\newblock The cityscapes dataset for semantic urban scene understanding.
\newblock In {\em Proceedings of the IEEE Conference on Computer Vision and
  Pattern Recognition (CVPR)}, 2016.

\bibitem{deng2009imagenet}
J. Deng, W. Dong, R. Socher, L.-J. Li, K. Li, and L. Fei-Fei.
\newblock Imagenet: A large-scale hierarchical image database.
\newblock In {\em Proceedings of the IEEE Conference on Computer Vision and
  Pattern Recognition (CVPR)}, 2009.

\bibitem{dhar2019learning_without_memorizing_gradcam}
Prithviraj Dhar, Rajat~Vikram Singh, Kuan-Chuan Peng, Ziyan Wu, and Rama
  Chellappa.
\newblock Learning without memorizing.
\newblock In {\em Proceedings of the IEEE Conference on Computer Vision and
  Pattern Recognition (CVPR)}, 2019.

\bibitem{douillard2020podnet}
Arthur Douillard, Matthieu Cord, Charles Ollion, Thomas Robert, and Eduardo
  Valle.
\newblock Podnet: Pooled outputs distillation for small-tasks incremental
  learning.
\newblock In {\em Proceedings of the IEEE European Conference on Computer
  Vision (ECCV)}, 2020.

\bibitem{douillard2020ghost}
Arthur Douillard, Eduardo Valle, Charles Ollion, and Matthieu Cord.
\newblock Insights from the future for continual learning.
\newblock In {\em arXiv preprint library}, 2020.

\bibitem{everingham2015pascalvoc}
Mark Everingham, S.~M. Ali~Eslami, Luc Van~Gool, Christopher K.~I. Williams,
  John~M. Winn, and Andrew Zisserman.
\newblock The pascal visual object classes challenge: {A} retrospective.
\newblock In {\em International Journal of Computer Vision (IJCV)}, 2015.

\bibitem{fernando2017path_net}
Chrisantha Fernando, Dylan Banarse, Charles Blundell, Yori Zwols, David Ha,
  Andrei~A. Rusu, Alexander Pritzel, and Daan Wierstra.
\newblock {PathNet: Evolution Channels Gradient Descent in Super Neural
  Networks}.
\newblock {\em arXiv preprint library}, 2017.

\bibitem{frankle2019lottery_ticket}
Jonathan Frankle and Michael Carbin.
\newblock The lottery ticket hypothesis: Finding sparse, trainable neural
  networks.
\newblock In {\em Proceedings of the International Conference on Learning
  Representations (ICLR)}, 2019.

\bibitem{french1999catastrophicforgetting}
Robert French.
\newblock Catastrophic forgetting in connectionist networks.
\newblock {\em Trends in cognitive sciences}, 1999.

\bibitem{fu2019DANet}
Jun Fu, Jing Liu, Haijie Tian, Yong Li, Yongjun Bao, Zhiwei Fang, and Hanqing
  Lu.
\newblock Dual attention network for scene segmentation.
\newblock In {\em Proceedings of the IEEE Conference on Computer Vision and
  Pattern Recognition (CVPR)}, 2019.

\bibitem{golkar2019neural_pruning}
Siavash Golkar, Michael Kagan, and Kyunghyun Cho.
\newblock Continual learning via neural pruning.
\newblock {\em Advances in Neural Information Processing Systems (NeurIPS)
  Workshop}, 2019.

\bibitem{hayes2020remind}
Tyler~L. Hayes, Kushal Kafle, Robik Shrestha, Manoj Acharya, and Christopher
  Kanan.
\newblock Remind your neural network to prevent catastrophic forgetting.
\newblock In {\em Proceedings of the IEEE European Conference on Computer
  Vision (ECCV)}, 2020.

\bibitem{he2014spatialpyramidpooling}
Kaiming He, Xiangyu Zhang, Shaoqing Ren, and Jian Sun.
\newblock Spatial pyramid pooling in deep convolutional networks for visual
  recognition.
\newblock In {\em Proceedings of the IEEE European Conference on Computer
  Vision (ECCV)}, 2014.

\bibitem{he2016resnet}
K. He, X. Zhang, S. Ren, and J. Sun.
\newblock Deep residual learning for image recognition.
\newblock In {\em Proceedings of the IEEE Conference on Computer Vision and
  Pattern Recognition (CVPR)}, 2016.

\bibitem{hinton2015knowledge_distillation}
Geoffrey Hinton, Oriol Vinyals, and Jeffrey Dean.
\newblock Distilling the knowledge in a neural network.
\newblock In {\em Advances in Neural Information Processing Systems (NeurIPS)
  Workshop}, 2015.

\bibitem{hou2020strippooling}
Qibin Hou, Li Zhang, Ming-Ming Cheng, and Jiashi Feng.
\newblock Strip pooling: Rethinking spatial pooling for scene parsing.
\newblock In {\em Proceedings of the IEEE Conference on Computer Vision and
  Pattern Recognition (CVPR)}, 2020.

\bibitem{hou2019ucir}
Saihui Hou, Xinyu Pan, Chen Change~Loy, Zilei Wang, and Dahua Lin.
\newblock Learning a unified classifier incrementally via rebalancing.
\newblock In {\em Proceedings of the IEEE Conference on Computer Vision and
  Pattern Recognition (CVPR)}, 2019.

\bibitem{huang2019CCNet}
Z. Huang, X. Wang, L. Huang, C. Huang, Y. Wei, and W. Liu.
\newblock Ccnet: Criss-cross attention for semantic segmentation.
\newblock In {\em Proceedings of the IEEE International Conference on Computer
  Vision (ICCV)}, 2019.

\bibitem{huang2020ccnet}
Zilong Huang, Xinggang Wang, Yunchao Wei, Lichao Huang, Humphrey Shi, Wenyu
  Liu, and Thomas~S. Huang.
\newblock Ccnet: Criss-cross attention for semantic segmentation.
\newblock 2020.

\bibitem{hung2019cpg}
Steven~C.Y. Hung, Cheng-Hao Tu, Cheng-En Wu, Chien-Hung Chen, Yi-Ming Chan, and
  Chu-Song Chen.
\newblock Compacting, picking and growing for unforgetting continual learning.
\newblock In {\em Advances in Neural Information Processing Systems (NeurIPS)},
  2019.

\bibitem{iscen2020incrementalfeatureadaptation}
Ahmet Iscen, Jeffrey Zhang, Svetlana Lazebnik, and Cordelia Schmid.
\newblock Memory-efficient incremental learning through feature adaptation.
\newblock In {\em Proceedings of the IEEE European Conference on Computer
  Vision (ECCV)}, 2020.

\bibitem{kato2019zeroshotsegmentation}
Naoki Kato, Toshihiko Yamasaki, and Kiyoharu Aizawa.
\newblock Zero-shot semantic segmentation via variational mapping.
\newblock In {\em Proceedings of the IEEE International Conference on Computer
  Vision (ICCV) Workshop}, 2019.

\bibitem{kemker2018fearnet}
Ronald Kemker and Christopher Kanan.
\newblock Fearnet: Brain-inspired model for incremental learning.
\newblock In {\em Proceedings of the International Conference on Learning
  Representations (ICLR)}, 2018.

\bibitem{kemker2018measuringforgetting}
Ronald Kemker, Marc McClure, Angelina Abitino, Tyler~L. Hayes, and Christopher
  Kanan.
\newblock Measuring catastrophic forgetting in neural networks.
\newblock In {\em Proceedings of the AAAI Conference on Artificial Intelligence
  (AAAI)}, 2018.

\bibitem{kim2019medic}
Dahyun Kim, Jihwan Bae, Yeonsik Jo, and Jonghyun Choi.
\newblock Incremental learning with maximum entropy regularization: Rethinking
  forgetting and intransigence.
\newblock {\em arXiv preprint library}, 2019.

\bibitem{kirkpatrick2017ewc}
James Kirkpatrick, Razvan Pascanu, Neil Rabinowitz, Joel Veness, Guillaume
  Desjardins, Andrei~A. Rusu, Kieran Milan, John Quan, Tiago Ramalho, Agnieszka
  Grabska-Barwinska, Demis Hassabis, Claudia Clopath, Dharshan Kumaran, and
  Raia Hadsell.
\newblock Overcoming catastrophic forgetting in neural networks.
\newblock {\em Proceedings of the National Academy of Sciences}, 2017.

\bibitem{kumar2018synthesized_zeroshot}
Vinay Kumar~Verma, Gundeep Arora, Ashish Mishra, and Piyush Rai.
\newblock Generalized zero-shot learning via synthesized examples.
\newblock In {\em Proceedings of the IEEE Conference on Computer Vision and
  Pattern Recognition (CVPR)}, 2018.

\bibitem{lampert2009zeroshot}
C.~H. Lampert, H. Nickisch, and S. Hermeling.
\newblock Learning to detect unseen object classes by between-class attribute
  transfer.
\newblock In {\em Proceedings of the IEEE Conference on Computer Vision and
  Pattern Recognition (CVPR)}, 2009.

\bibitem{lazbnik2006spatial_pyramid_matching}
Svetlana Lazebnik, Cordelia Schmid, and Jean Ponce.
\newblock Beyond bags of features: Spatial pyramid matching for recognizing
  natural scene categories.
\newblock {\em Object Categorization: Computer and Human Vision Perspectives,
  Cambridge University Press}, 2006.

\bibitem{lee2013pseudolabel}
Dong-Hyun Lee.
\newblock Pseudo-label: The simple and efficient semi-supervised learning
  method for deep neural networks.
\newblock In {\em International Conference on Machine Learning (ICML)
  Workshop}, 2013.

\bibitem{li2019learning_to_grow}
Xilai Li, Yingbo Zhou, Tianfu Wu, Richard Socher, and Caiming Xiong.
\newblock Learn to grow: {A} continual structure learning framework for
  overcoming catastrophic forgetting.
\newblock {\em Proceedings of the International Conference on Learning
  Representations (ICLR)}, 2019.

\bibitem{li2019bidirectionallearning}
Yunsheng Li, Lu Yuan, and Nuno Vasconcelos.
\newblock Bidirectional learning for domain adaptation of semantic
  segmentation.
\newblock In {\em Proceedings of the IEEE Conference on Computer Vision and
  Pattern Recognition (CVPR)}, 2019.

\bibitem{li2018lwf}
Z. Li and D. Hoiem.
\newblock Learning without forgetting.
\newblock {\em Proceedings of the IEEE European Conference on Computer Vision
  (ECCV)}, 2016.

\bibitem{liu2020mnemonics}
Yaoyao Liu, Yuting Su, An-An Liu, Bernt Schiele, and Qianru Sun.
\newblock Mnemonics training: Multi-class incremental learning without
  forgetting.
\newblock In {\em Proceedings of the IEEE Conference on Computer Vision and
  Pattern Recognition (CVPR)}, 2020.

\bibitem{lomonaco2017core50}
Vincenzo Lomonaco and Davide Maltoni.
\newblock Core50: a new dataset and benchmark for continuous object
  recognition.
\newblock In {\em Annual Conference on Robot Learning}, 2017.

\bibitem{lomonaco2020ar1}
Vincenzo Lomonaco, Davide Maltoni, and Lorenzo Pellegrini.
\newblock Rehearsal-free continual learning over small non-i.i.d. batches.
\newblock In {\em Proceedings of the IEEE Conference on Computer Vision and
  Pattern Recognition (CVPR) Workshop}, 2020.

\bibitem{long2015fcn}
J. {Long}, E. {Shelhamer}, and T. {Darrell}.
\newblock Fully convolutional networks for semantic segmentation.
\newblock In {\em Proceedings of the IEEE Conference on Computer Vision and
  Pattern Recognition (CVPR)}, 2015.

\bibitem{lopezpaz2017gem}
David Lopez-Paz and Marc'Aurelio Ranzato.
\newblock Gradient episodic memory for continual learning.
\newblock In I. Guyon, U.~V. Luxburg, S. Bengio, H. Wallach, R. Fergus, S.
  Vishwanathan, and R. Garnett, editors, {\em Advances in Neural Information
  Processing Systems (NeurIPS)}, 2017.

\bibitem{mehta2018espnet}
Sachin Mehta, Mohammad Rastegari, Anat Caspi, Linda~G. Shapiro, and Hannaneh
  Hajishirzi.
\newblock Efficient spatial pyramid of dilated convolutions for semantic
  segmentation.
\newblock In {\em Proceedings of the IEEE European Conference on Computer
  Vision (ECCV)}, 2018.

\bibitem{michieli2019ilt}
Umberto Michieli and Pietro Zanuttigh.
\newblock Incremental learning techniques for semantic segmentation.
\newblock In {\em Proceedings of the IEEE International Conference on Computer
  Vision (ICCV) Workshop}, 2019.

\bibitem{noh2015deconvolution}
H. Noh, S. Hong, and B. Han.
\newblock Learning deconvolution network for semantic segmentation.
\newblock In {\em Proceedings of the IEEE International Conference on Computer
  Vision (ICCV)}, 2015.

\bibitem{ozdemir2018learnthenewkeeptheold}
Firat Ozdemir, Philipp Fuernstahl, and Orcun Goksel.
\newblock Learn the new, keep the old: Extending pretrained models with new
  anatomy and images.
\newblock In {\em International Conference on Med- ical Image Computing and
  Computer-Assisted Intervention}, 2018.

\bibitem{ozdemir2019segmentationanotomical}
Firat Ozdemir and Orcun Goksel.
\newblock Extending pretrained segmentation networks with additional anatomical
  structures.
\newblock In {\em International journal of computer assisted radiology and
  surgery}, 2019.

\bibitem{park2020csc}
Sangyong Park and Yong~Seok Heo.
\newblock Knowledge distillation for semantic segmentation using channel and
  spatial correlations and adaptive cross entropy.
\newblock In {\em Sensors}, 2020.

\bibitem{paszke2017pytorch}
Adam Paszke, Sam Gross, Soumith Chintala, Gregory Chanan, Edward Yang, Zachary
  DeVito, Zeming Lin, Alban Desmaison, Luca Antiga, and Adam Lerer.
\newblock Automatic differentiation in pytorch.
\newblock In {\em Advances in Neural Information Processing Systems (NeurIPS)
  Workshop}, 2017.

\bibitem{rebuffi2017icarl}
Sylvestre-Alvise Rebuffi, Alexander Kolesnikov, Georg Sperl, and Christoph~H.
  Lampert.
\newblock icarl: Incremental classifier and representation learning.
\newblock In {\em Proceedings of the IEEE Conference on Computer Vision and
  Pattern Recognition (CVPR)}, 2017.

\bibitem{robins1995catastrophicforgetting}
Anthony Robins.
\newblock Catastrophic forgetting, rehearsal and pseudorehearsal.
\newblock {\em Connection Science}, 1995.

\bibitem{romero2014fitnet_hints}
Adriana Romero, Nicolas Ballas, Samira~Ebrahimi Kahou, Antoine Chassang, Carlo
  Gatta, and Yoshua Bengio.
\newblock Fitnets: Hints for thin deep nets.
\newblock {\em arXiv preprint library}, 2014.

\bibitem{ronneberger2015UNet}
Olaf Ronneberger, Philipp Fischer, and Thomas Brox.
\newblock U-net: Convolutional networks for biomedical image segmentation.
\newblock In {\em International Conference on Medical Image Computing and
  Computer Assisted Intervention (MICCAI)}, 2015.

\bibitem{saporta2020esl}
Antoine Saporta, Tuan-Hung Vu, Matthieu Cord, and Patrick Pérez.
\newblock Esl: Entropy-guided self-supervised learning for domain adaptation in
  semantic segmentation.
\newblock In {\em Proceedings of the IEEE Conference on Computer Vision and
  Pattern Recognition (CVPR) Workshop}, 2020.

\bibitem{sermanet2014overfeat}
Pierre Sermanet, David Eigen, Xiang Zhang, Micha{\"{e}} Mathieu, Rob Fergus,
  and Yann LeCun.
\newblock Overfeat: Integrated recognition, localization and detection using
  convolutional networks.
\newblock In {\em Proceedings of the International Conference on Learning
  Representations (ICLR)}, 2014.

\bibitem{shin2017deep_generative_replay}
Hanul Shin, Jung~Kwon Lee, Jaehong Kim, and Jiwon Kim.
\newblock Continual learning with deep generative replay.
\newblock In {\em Advances in Neural Information Processing Systems (NeurIPS)},
  2017.

\bibitem{tao2020HRNet}
Andrew Tao, Karan Sapra, and Bryan Catanzaro.
\newblock Hierarchical multi-scale attention for semantic segmentation.
\newblock In {\em arXiv preprint library}, 2020.

\bibitem{thrun1998lifelonglearning}
Sebastian Thrun.
\newblock Lifelong learning algorithms.
\newblock In {\em Springer Learning to Learn}, 1998.

\bibitem{vu2019advent}
Tuan-Hung Vu, Himalaya Jain, Maxime Bucher, Matthieu Cord, and Patrick Pérez.
\newblock Advent: Adversarial entropy minimization for domain adaptation in
  semantic segmentation.
\newblock In {\em Proceedings of the IEEE Conference on Computer Vision and
  Pattern Recognition (CVPR)}, 2019.

\bibitem{wang2020axialdeeplab}
Huiyu Wang, Yukun Zhu, Bradley Green, Hartwig Adam, Alan Yuille, and
  Liang-Chieh Chen.
\newblock Axial-deeplab: Stand-alone axial-attention for panoptic segmentation.
\newblock In {\em Proceedings of the IEEE European Conference on Computer
  Vision (ECCV)}, 2020.

\bibitem{wang2020bookworm}
Kai Wang, Luis Herranz, Anjan Dutta, and Joost van~de Weijer.
\newblock Bookworm continual learning: beyond zero-shot learning and continual
  learning.
\newblock In {\em Proceedings of the IEEE European Conference on Computer
  Vision (ECCV) Workshop}, 2020.

\bibitem{wortsman2020supermasks}
Mitchell Wortsman, Vivek Ramanujan, Rosanne Liu, Aniruddha Kembhavi, Mohammad
  Rastegari, Jason Yosinski, and Ali Farhadi.
\newblock Supermasks in superposition for continual learning.
\newblock In {\em Advances in Neural Information Processing Systems (NeurIPS)},
  2020.

\bibitem{wu2019bias_correction}
Yue Wu, Yinpeng Chen, Lijuan Wang, Yuancheng Ye, Zicheng Liu, Yandong Guo, and
  Yun Fu.
\newblock Large scale incremental learning.
\newblock In {\em Proceedings of the IEEE Conference on Computer Vision and
  Pattern Recognition (CVPR)}, 2019.

\bibitem{yoon2018dynamically_expandable_networks}
Jaehong Yoon, Eunho Yang, Jeongtae Lee, and Sung~Ju Hwang.
\newblock Lifelong learning with dynamically expandable networks.
\newblock In {\em Proceedings of the International Conference on Learning
  Representations (ICLR)}, 2018.

\bibitem{yuan2020ocr}
Yuhui Yuan, Xilin Chen, and Jingdong Wang.
\newblock Object-contextual representations for semantic segmentation.
\newblock In {\em Proceedings of the IEEE European Conference on Computer
  Vision (ECCV)}, 2020.

\bibitem{yuan2018ocnet}
Yuhui Yuan and Jingdong Wang.
\newblock Ocnet: Object context network for scene parsing.
\newblock In {\em arXiv preprint library}, 2018.

\bibitem{zagoruyko2016distillation_attention}
Sergey Zagoruyko and Nikos Komodakis.
\newblock Paying more attention to attention: Improving the performance of
  convolutional neural networks via attention transfer.
\newblock {\em Proceedings of the International Conference on Learning
  Representations (ICLR)}, 2016.

\bibitem{zenke2017synaptic_intelligence}
Friedemann Zenke, Ben Poole, and Surya Ganguli.
\newblock Continual learning through synaptic intelligence.
\newblock In {\em International Conference on Machine Learning (ICML)}, 2017.

\bibitem{zhang2018ContextEncoding}
Hang Zhang, Kristin~J. Dana, Jianping Shi, Zhongyue Zhang, Xiaogang Wang,
  Ambrish Tyagi, and Amit Agrawal.
\newblock Context encoding for semantic segmentation.
\newblock In {\em Proceedings of the IEEE Conference on Computer Vision and
  Pattern Recognition (CVPR)}, 2018.

\bibitem{zhang2020resnest}
Hang Zhang, Chongruo Wu, Zhongyue Zhang, Yi Zhu, Zhi Zhang, Haibin Lin, Yue
  Sun, Tong He, Jonas Muller, R. Manmatha, Mu Li, and Alexander Smola.
\newblock Resnest: Split-attention networks.
\newblock In {\em arXiv preprint library}, 2020.

\bibitem{zhao2020weightalignement}
Bowen Zhao, Xi Xiao, Guojun Gan, Bin Zhang, and Shutao Xia.
\newblock Maintaining discrimination and fairness in class incremental
  learning.
\newblock In {\em Proceedings of the IEEE Conference on Computer Vision and
  Pattern Recognition (CVPR)}, 2020.

\bibitem{zhao2017PSPNet}
H. Zhao, J. Shi, X. Qi, X. Wang, and J. Jia.
\newblock Pyramid scene parsing network.
\newblock In {\em Proceedings of the IEEE Conference on Computer Vision and
  Pattern Recognition (CVPR)}, 2017.

\bibitem{zhao2018psanet}
Hengshuang Zhao, Yi Zhang, Shu Liu, Jianping Shi, Chen~Change Loy, Dahua Lin,
  and Jiaya Jia.
\newblock Psanet: Point-wise spatial attention network for scene parsing.
\newblock In {\em Proceedings of the IEEE European Conference on Computer
  Vision (ECCV)}, 2018.

\bibitem{zhou2017adedataset}
Bolei Zhou, Hang Zhao, Xavier Puig, Sanja Fidler, Adela Barriuso, and Antonio
  Torralba.
\newblock Scene parsing through ade20k dataset.
\newblock In {\em Proceedings of the IEEE Conference on Computer Vision and
  Pattern Recognition (CVPR)}, 2017.

\bibitem{peng2019m2kd}
Peng Zhou, Long Mai, Jianming Zhang, Ning Xu, Zuxuan Wu, and Larry~S. Davis.
\newblock M2kd: Multi-model and multi-level knowledge distillation for
  incremental learning.
\newblock {\em arXiv preprint library}, 2019.

\bibitem{zou2018classbalancedselftraining}
Yang Zou, Zhiding Yu, BVK Vijaya~Kumar, and Jinsong Wang.
\newblock Unsupervised domain adaptation for semantic seg- mentation via
  class-balanced self-training.
\newblock In {\em Proceedings of the IEEE European Conference on Computer
  Vision (ECCV)}, 2018.

\end{thebibliography}
}

\end{document}